\documentclass{article}

\usepackage{PRIMEarxiv}

\usepackage[utf8]{inputenc} 
\usepackage[T1]{fontenc}    
\usepackage{hyperref}       
\usepackage{url}            
\usepackage{booktabs}       
\usepackage{amsfonts}       
\usepackage{nicefrac}       
\usepackage{microtype}      
\usepackage{lipsum}
\usepackage{fancyhdr}       
\usepackage{graphicx}       
\graphicspath{{media/}}     
\usepackage{tabularx}
\usepackage{mathtools}
\usepackage{amssymb}
\usepackage{multirow}
\usepackage{subcaption}
\title{Counterfactual Explanations for Deep Learning-Based Traffic Forecasting}

\author{
  Rushan Wang\textsuperscript{a,b}, Yanan Xin\textsuperscript{a}, Yatao Zhang\textsuperscript{a}, Fernando Perez-Cruz\textsuperscript{c}, Martin Raubal\textsuperscript{a} \\
  \textsuperscript{a}Institute of Cartography and Geoinformation, ETH Zurich, Switzerland \\
  \textsuperscript{b}WSL Institute for Snow and Avalanche Research SLF, Davos Dorf, Switzerland \\
  \textsuperscript{a}Institute for Machine Learning, Department of Computer Science, ETH Zurich, Switzerland \\
}

\begin{document}
\maketitle

\begin{abstract}
Deep learning models are widely used in traffic forecasting and have achieved state-of-the-art prediction accuracy. However, the black-box nature of those models makes the results difficult to interpret by users. 
This study aims to leverage an Explainable AI approach, counterfactual explanations, to enhance the explainability and usability of deep learning-based traffic forecasting models. 
Specifically, the goal is to elucidate relationships between various input contextual features and their corresponding predictions. 
We present a comprehensive framework that generates counterfactual explanations for traffic forecasting and provides usable insights through the proposed scenario-driven counterfactual explanations. 
The study first implements a deep learning model to predict traffic speed based on historical traffic data and contextual variables. Counterfactual explanations are then used to illuminate how alterations in these input variables affect predicted outcomes, thereby enhancing the transparency of the deep learning model.
We investigated the impact of contextual features on traffic speed prediction under varying spatial and temporal conditions. The scenario-driven counterfactual explanations integrate two types of user-defined constraints, directional and weighting constraints, to tailor the search for counterfactual explanations to specific use cases. These tailored explanations benefit machine learning practitioners who aim to understand the model's learning mechanisms and domain experts who seek insights for real-world applications.
Our findings underscore the integral relationship between traffic speed prediction and diverse contextual features, displaying varied patterns across suburban and urban roads, as well as weekdays and weekends. 
The results showcase the effectiveness of counterfactual explanations in revealing traffic patterns learned by deep learning models, showing its potential for interpreting black-box deep learning models used for spatiotemporal predictions in general. 
\end{abstract}


\section{Introduction}
\label{introduction}

Accurate traffic forecasting is integral to building Intelligent Transportation Systems, which can help alleviate traffic congestion, improve traffic operation efficiency, and reduce carbon emissions \cite{its}.
Research on traffic forecasting has focused on capturing the temporal and spatial dependencies in traffic data and predicting dynamic traffic states such as traffic flow, traffic speed, and traffic demand.
Over the last few years, the focus of traffic forecasting methods has shifted from using classical statistical techniques \cite{lee1999application, wu2004travel, zarei2013road} to data-driven machine/deep learning methods such as Recurrent Neural Network, Long Short-Term Memory, or Graph Neural Network \cite{polson2017deep}. 
The performance of traffic forecasting benefited significantly from the advancement of deep learning techniques and artificial intelligence \cite{Yin_2022}. A considerable number of studies have demonstrated the exceptional performance of deep learning algorithms in reducing predictive errors in traffic forecasting.
However, challenges arise with the black-box nature of these deep learning models. The lack of interpretability and explainability makes it difficult for machine learning developers to understand the learning mechanisms of these models \cite{xin2023evaluating}. Furthermore, it is also challenging for domain experts to utilize these models and derive insightful understandings of traffic dynamics due to the opacity of the models \cite{jonietz2022urban}.
These challenges hinder the adoption of deep learning models in practice \cite{Fernandez2020ExplainingDD}. 

Recently, the issues of interpretability and explainability in AI gained increasing attention from researchers \cite{article_explainabilityinterpretability}.
To address this challenge, Explainable Artificial Intelligence (XAI) techniques are proposed to enhance ML models' interpretability and explainability, making the output of these models more comprehensible to humans \cite{edwards2017slave}.
One type of commonly used XAI method is local explanations, which involve using a simpler surrogate model to approximate the decisions of the model at a local region to yield interpretable information, for example, feature importance scores \cite{lundberg2017unified}. However, these techniques suffer from an inherent fidelity-interpretability trade-off due to the use of a simpler model for generating explanations. 
On the contrary, Counterfactual Explanations (CFEs) as a local explanation method can maintain consistency with the original machine learning model, offering insights into the inner workings of machine learning models \cite{wachter2018counterfactual}. 
CFEs reveal the minimal changes required in the original input features to alter the model's prediction, thus providing understanding without sacrificing fidelity or complexity. 

In our study, CFEs are particularly advantageous since we are interested in determining the minimal change in the input to obtain a desired alternative prediction. CFEs are straightforward to understand and can be used to provide users with a course of action to alter the prediction if they receive unfavourable decisions. These explanations establish a relationship between the input features and the decision, making them highly valuable for users to comprehend, interact with, and utilize these models.

Currently, there is a significant lack of study in applying XAI techniques in the domain of traffic forecasting, or in general spatiotemporal data analysis \cite{xin2022vision, li4370154interpreting, yang2023counterfactual}. It is not straightforward to apply counterfactual methods developed in non-spatial domains to spatiotemporal data analysis due to the high complexity and dimensionality of spatiotemporal data \cite{xin2022vision}. Thus, one of the core objectives of this study is to explore the potential and limitations of counterfactual explanations in deep learning-based traffic forecasting applications.

The study is guided by the following research questions:
\begin{itemize}
    \item What is the impact of input variables on deep learning-based traffic forecasting?
    \item How can we modify the input variables to achieve the desired prediction for various scenarios?
\end{itemize}

This paper involves training and explaining a deep-learning model for traffic forecasting. Particularly, by applying the XAI technique, the study contributes to our understanding of how the model produces predictions, and how variations in input features can affect predicted results.
The second key contribution of our study is the application of CFEs on spatiotemporal prediction tasks, where the spatiotemporal dependencies are critical. In this context, we conduct a thorough evaluation of the impact of the counterfactual features on the spatiotemporal traffic dynamics.
Another contribution of this study is the proposal of scenario-driven counterfactual explanations, where we demonstrate and validate different methods to integrate user prior knowledge or constraints in generating counterfactuals.

In summary, this study proposes a framework to tackle the lack of explainability of black-box traffic forecasting models. By streamlining the procedures of generating and examining counterfactual explanations in deep learning-based traffic forecasting, this study offers valuable insights for future studies in this direction.

\section{Related Work}

\subsection{Deep Learning in traffic forecasting} \label{relateworktraffic}

It is an important research topic to analyze the non-linear and complex spatiotemporal patterns of traffic dynamics in order to make accurate traffic predictions \cite{Yin_2022}. 
Statistical and traditional machine learning models are two major representative data-driven methods for traffic prediction. This includes methods such as Historical Average (HA), Auto-Regressive Integrated Moving Average (ARIMA) \cite{statisticmethod}, Support Vector Regression \cite{SVM}, and Random Forest Regression \cite{rfr}.
However, one of the disadvantages of traditional approaches is that most of the applied features need to be carefully selected and processed by a domain expert to reduce the complexity of the feature space and make the underlying patterns easier to extract.

Over the last few years, deep learning-based methods have unlocked the potential of artificial intelligence in traffic prediction \cite{Lv2015TrafficFP}.
Deep learning models exploit much more features and complex architectures than classical methods and can achieve better performance. Recurrent Neural Networks (RNNs) stand out as particularly effective in time series forecasting \cite{prasad2014deep, rnntraffic}.
Additionally, a series of studies have applied CNN to capture spatial correlations in traffic networks from two-dimensional spatiotemporal traffic data \cite{articleCNN}. However, the CNN-based approach is not optimal for traffic foresting problems that have a graph-based data type.

Over the past few years, graph neural networks (GNNs) have emerged as a cutting-edge deep learning technique, demonstrating state-of-the-art performance in numerous applications \cite{GNNWu}.
Due to their capability of modeling non-Euclidean graph structures, GNNs are particularly well-suited for traffic forecasting tasks where complex spatial dependencies need to be captured \cite{WeiGNN}.
These include, for instance, the diffusion convolutional recurrent neural network (DCRNN) \cite{DCRNN}, temporal graph convolutional network (T-GCN) \cite{TGCN}, and Graph WaveNet \cite{graphwavenet} models.

In traffic prediction studies, contextual data has been widely recognized as an important input to improve traffic prediction performance \cite{zhang2023incorporating}. Some commonly used external variables include weather conditions, events, and time information \cite{Yin_2022}.
One previous study \cite{liao2018deep} incorporated auxiliary data, such as crowd map queries and road intersections, along with geographical and social variables, into an encoder-decoder sequence learning framework for traffic forecasting. In another study \cite{zhu2020astgcn}, researchers categorized these influencing factors as either dynamic or static attributes and designed an attribute-augmented unit that seamlessly integrates these variables into a spatiotemporal graph convolution model, which enhanced the model's forecasting capabilities. Classifying contextual data into spatial and temporal contextual features, \cite{zhang2023incorporating} proposed a multimodal context-based graph convolutional neural network (MCGCN) to embed spatial and temporal contexts and incorporate them into traffic speed prediction for better performance.

\subsection{Counterfactual Explanations}
Counterfactual Explanations (CFEs) suggest what should be different in the input instance to change the outcome of an AI system \cite{wachter2018counterfactual, Lucic_2020}. 
In recent years, CFEs have been applied in various tasks to enhance the interpretability of machine/deep learning models \cite{Fernandez2020ExplainingDD}. 
It has already been widely used in image classification, where generative models such as GANs and variational autoencoders (VAE) are used to implement interventions and generate realistic CFEs \cite{image1, image2, image3,dlwithCFE}.
Other than image data, CFEs have also been utilized for text data \cite{Jung_2022}, speech data \cite{Zhang_2022}, time-series data \cite{Ates_2021}, and graph data \cite{CFEGNN}, etc. 

Numerous methods are developed for generating CFEs, each with its specific focus and application. For instance, the FACE method \cite{Poyiadzi_2020} aims to produce plausible CFEs by building feasible paths between data points associated with opposing predictions. On the other hand, DiCE \cite{DICE} is designed primarily for differentiable models and is especially useful for handling continuous features. Another innovative approach is the Bayesian-optimization-based Counterfactual Explanations \cite{bayes}, which employ probabilistic methods to generate counterfactuals. Additionally, Multi-Objective Counterfactuals (MOC) \cite{Dandl_2020} was proposed recently that conceptualizes the counterfactual search as a multi-objective optimization problem, which broadens the scope and applicability of CFEs in complex scenarios. In this study, we used MOC due to its ability to produce a varied set of counterfactuals, offering multiple options for actionable feature adjustments based on different objective trade-offs.

\section{Methods}

\subsection{Data}

The traffic speed data was provided by HERE technologies\footnote{HERE Technologies, URL: \url{https://developer.here.com/products/platform/data}}, which offers a record of traffic speed observations on different road segments. 

\begin{figure}[ht]
    \centering
    \includegraphics[width=0.6\textwidth]{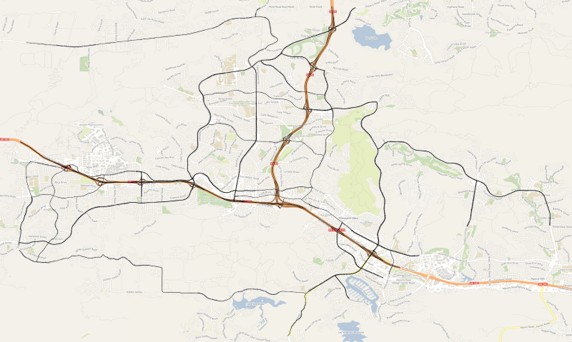}
    \caption{Location of road network (dark line).}
    \label{fig:road}
\end{figure}

\begin{figure}[htb]
    \centering
    \includegraphics[width=0.8\textwidth]{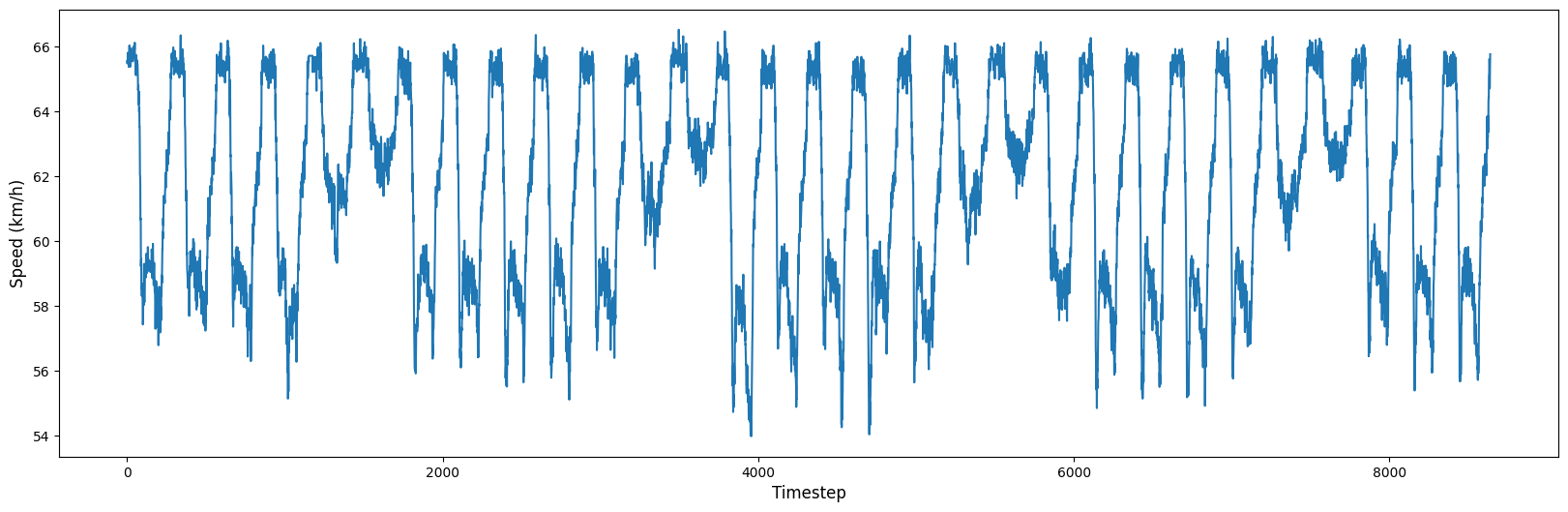}
    \caption{Average speed for all the 3169 road segments from January 1st to 30th, 2019.}
    \label{fig:averageoverall}
\end{figure}

In this study, the road graph is located in Thousand Oaks, California, USA, as shown in Figure \ref{fig:road}, which consists of 3169 road segments.
The data were collected from January 1st to January 30th, 2019, at 5-minute intervals.
Figure \ref{fig:averageoverall} shows the average speed of all the road segments within the study period. A noticeable temporal pattern emerges, where lower speeds appear during the daytime and a distinct weekly pattern exists with different speed variations between weekdays and weekends.

Contextual data is of great importance to traffic prediction. 
In this study, several contextual features were collected, which can be classified into static features and dynamic features. 
Static features are location-based, which vary with regard to different road segments. 
Based on findings from previous studies (see section \ref{relateworktraffic}), this study included nearby POI data, speed limit data, and lane configuration of each road segment as static features. 
Particularly, the POIs include the nearby gas station, charging station, parking lot, and restaurant. 
Dynamic features are time-based features that change over time. In our study, dynamic features such as the day of the week, hour of the day, and weather condition data (e.g., temperature, wind speed, precipitation, humidity) are included.
Table \ref{contexttable} summarizes all the contextual features involved in the study.

\begin{table}[htb]
    \centering
    \begin{tabular}{c|c|c}
         \toprule
   Class & Contextual data & Encoding method \\
   \midrule
   \multirow{3}{*}{Static feature} & Number of POIs & Integer\\
                                   & Speed limit & Integer\\
                                   & Number of lanes & Integer\\
   \midrule
   \multirow{6}{*}{Dynamic feature} & Day of the week & One-hot encoding\\
                                    & Hour of the day & Sin-cos encoding\\
                                    & Temperature & Float\\
                                    & Wind speed & Float\\
                                    & Precipitation & Float\\
                                    & Humidity & Float\\
   \bottomrule
    \end{tabular}
    \caption{Summary of the contextual data in this study and their encoding method.}
    \label{contexttable}
\end{table}

\subsection{Traffic forecasting model}
In this study, the traffic forecasting model is built to predict the future traffic speed for each road segment of the traffic graph. Specifically, the definitions of traffic graph and graph-based traffic forecasting are as follows:

\begin{itemize}
    \item Traffic Graph:
     A graph G = (V, E, A) can be utilized to describe the topological structure of the road network, and each road segment is treated as a node, where V is a set of road nodes, \(V = \{v_1,v_2,\ldots,v_N\}\), N is the number of the nodes, and E is a set of edges. The adjacency matrix A is used to represent connections between road segments, \(A\in R_{N×N}\). 
    \item Graph-Based Traffic Forecasting:
    The spatiotemporal traffic forecasting task can be defined as to find a function \(f\) which generates \(y = f(\chi,\varepsilon ;G)\), where \(y\) is the traffic state to be predicted, \(\chi = \{\chi_1, \chi_2, \ldots, \chi _T\}\) is the historical traffic state defined on graph G, T is the number of time steps in the historical window size, and \(\varepsilon\) represents the external factors.     
\end{itemize}

\begin{figure}[htb]
    \centering
    \includegraphics[width=0.75\textwidth]{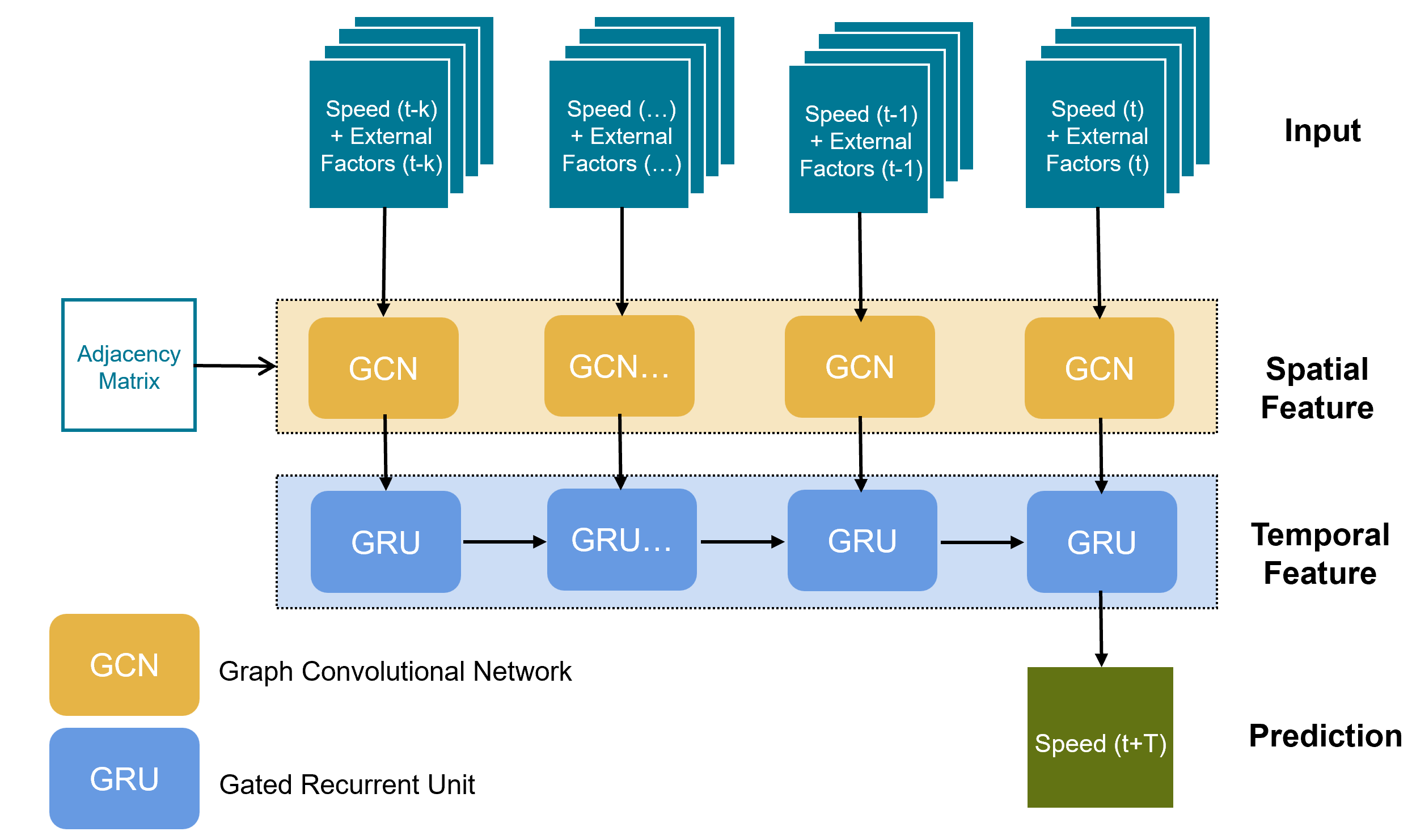}
    \caption{The architecture of the deep learning model used in this study for traffic forecasting.}
    \label{fig:astgcn}
\end{figure}

Inspired by the temporal graph convolutional network model \cite{TGCN} and AST-GCN model \cite{zhu2020astgcn}, this study adopted a similar model. 
Figure \ref{fig:astgcn} shows the architecture of the deep learning model we used.

For each input unit at time step \(t\), traffic speed data \(\chi_t\) and contextual data \(\varepsilon_t\) are concatenated as enhanced feature matrix \(X_t\). Together with adjacency matrix \(A\), they are fed into the graph convolutional network (GCN), which can capture the spatial dependence of the data. The modelling process of GCN can be expressed as \cite{zhu2020astgcn}:

\begin{equation}
    gc_{l+1}=\sigma(\widetilde{D}^{-\frac{1}{2}} \widetilde{A} \widetilde{D}^{-\frac{1}{2}} gc_l W_l)
\end{equation}

where \(\sigma\) is the activation function, \(\widetilde{A} = A+I\) represents a matrix with self-connection structure, \(\widetilde{D}\) is a degree matrix, \(W_l\) denotes the weight matrix of the \(l\)-th convolutional layer, \(c_l\) is the output representation, and \(gc_0=X\), \(X\) is the feature matrix.

To capture the temporal features, the architecture combines GCN and GRU models. Specifically, the feature matrics are fed into a series of GCNs to generate time-varying features. Then the feature series are used as input of GRUs to model the temporal dependence and derive hidden traffic states.

\begin{figure}[htb]
    \centering
    \includegraphics[width=0.75\textwidth]{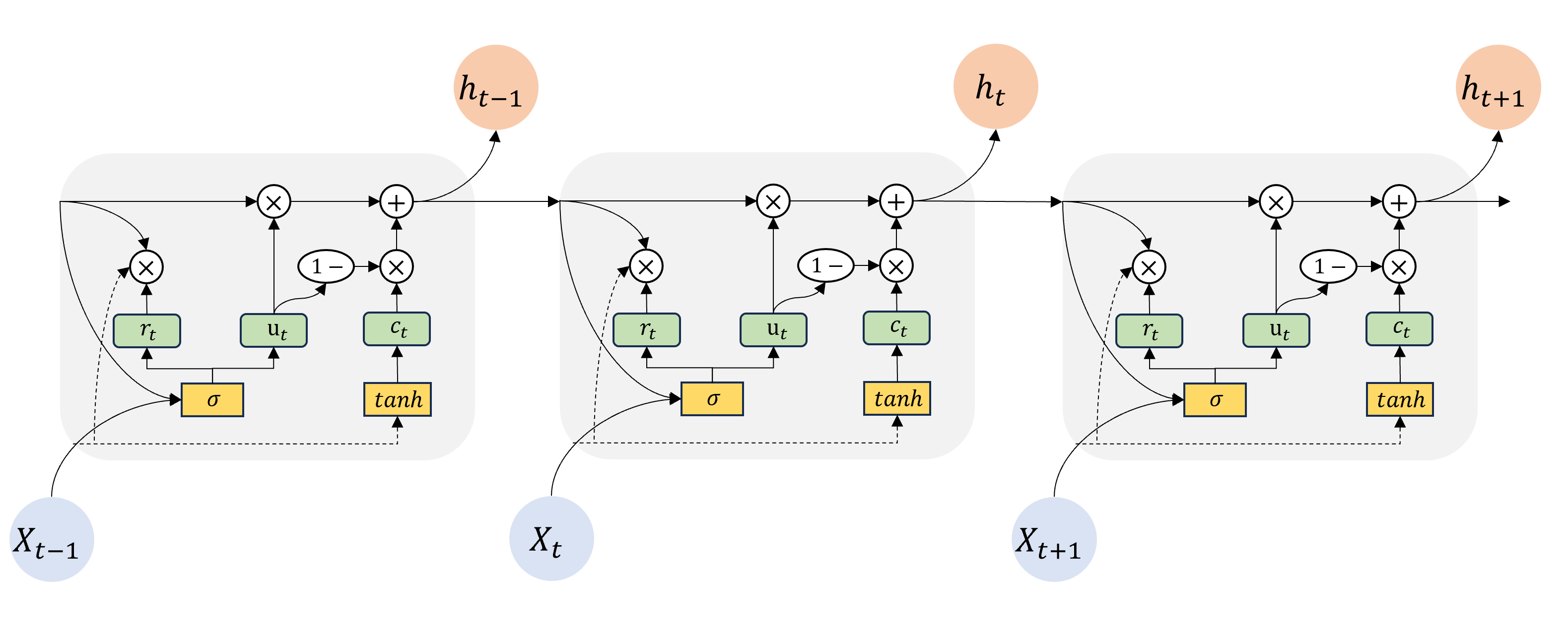}
    \caption{The architecture of the Gated Recurrent Unit (GRU) model \cite{zhu2020astgcn}.}
    \label{fig:gru}
\end{figure}

As shown in Figure \ref{fig:gru}, \(h_{t-1}\)denotes the output at time \(t-1\), \(gc\) is graph convolution process, \(u_t\) and \(r_t\) are update gate and reset gate at time \(t\), and \(h_t\) denotes the output at time \(t\). The specific calculation process is shown below, where \(W\) and \(b\) are the weights and deviations in the training process:

\begin{equation}
    u_t = \sigma(W_u\cdot [gc(X_t, A), h_{t-1}]+b_u)
\end{equation}
\begin{equation}
    r_t = \sigma(W_r\cdot [gc(X_t, A), h_{t-1}]+b_r)
\end{equation}\begin{equation}
    c_t = tanh(W_c\cdot [gc(X_t, A), (r_t, h_{t-1})]+b_c)
\end{equation}\begin{equation}
    h_t = u_t \ast h_{t-1} + (1-u_t) \ast c_t
\end{equation}

During the training process, the loss function is set to minimize the variation between the real traffic speed and the predicted speed.

\begin{equation}
    Loss = \left \| y_t - \hat{y_t} \right \| +\lambda L_{reg}
\end{equation}

where \(y_t\) and \(\hat{y_t}\) are the ground truth and prediction, \(L_{reg}\) is the L1 regularisation term to avoid overfitting, and \(\lambda\) is a hyperparameter.

The following metrics were used to evaluate the prediction accuracy of the model:  

\begin{itemize}
    \item Root Mean Squared Error (RMSE):
    \begin{equation}
    RMSE = \sqrt{\frac{1}{n}\sum_{t=1}^{n} (y_t - \hat{y_t})^{2}}
\end{equation}
    \item Mean absolute error (MAE):
    \begin{equation}
        MAE = \frac{1}{n}\sum_{t=1}^{n} \left |y_t - \hat{y_t} \right |
    \end{equation}
    \item Accuracy
    \begin{equation}
        Accuracy = 1 - \frac{\left \| y - \hat{y} \right \|_{F}}{\left \| y \right \|_F}
    \end{equation}
    where \(\left \| \cdot  \right \|_F\) is the Frobenius norm.
    \item Coefficient of Determination (\(R^2\))
    \begin{equation}
        R^2 = 1-\frac{\sum_{t = 1}^{}(y_t - \hat{y_t})}{\sum_{t = 1}^{}(y_t - \bar{y_t})}
    \end{equation}
    \item Explained variation (VAR)
    \begin{equation}
        Var = 1 - \frac{Var(y - \hat{y})}{Var(y)}
    \end{equation}
    This measures the proportion to which the proposed model accounts for the variation in real traffic states, which is mainly used to measure the predictive ability of the model.
\end{itemize}

\subsection{Multi-objective optimization to select CFEs} \label{MOC_objective}

When generating CFEs, there can be multiple possibilities to conduct changes in input features to achieve the desired alternative prediction. Therefore, different criteria or objectives are proposed to help select optimal CFEs.
Existing approaches to generate counterfactual explanations often rely on optimizing a single weighted sum of multiple objectives, making it difficult to balance different objectives. Following the approach proposed by \cite{Dandl_2020}, this study considers the task of generating counterfactual explanations as a multi-objective optimization problem, which allows for the generation of a diverse set of CFEs.

Multi-objective optimization is a mathematical technique used for solving problems involving competing objectives. In the context of counterfactual explanations, the goal is to optimize for multiple criteria simultaneously, rather than aggregating them into a single metric.

To guide the search for counterfactuals, we employed four key criteria, which are:

\begin{itemize}
    \item \textbf{Validity:} A counterfactual is valid if it produces a predicted outcome closely approximating the target speed.
    \item \textbf{Proximity:} The ideal counterfactual should differ minimally from the original feature set, thereby ensuring that the changes suggested are modest and realistic.
    \item \textbf{Sparsity:} A counterfactual gains in feasibility when the number of altered features is minimized.
    \item \textbf{Plausibility:} For a counterfactual explanation to be considered plausible, it should be close to the nearest observed data points.
\end{itemize}

It is important to recognize that a counterfactual example, while perhaps optimal in feature space, may not be practically feasible due to real-world constraints. Therefore, users should also have the flexibility to specify constraints on specific features, including:

\begin{itemize}
    \item \textbf{Range Constraints:} These define feasible ranges for each feature. For instance, a constraint might specify that ``Speed limit on the road should be larger than 30 km/h."
    \item \textbf{Mutable Variables:} Alternatively, users may specify which variable can be altered in the search for a counterfactual explanation. 
\end{itemize}

The presence of multiple objectives in a problem gives rise to a set of optimal solutions, known as Pareto-optimal solutions. Without additional information, it is hard to say which Pareto-optimal solution is better than the others. 
To efficiently address this problem, we used the Non-dominated Sorting Genetic Algorithm II (NSGA-II), a fast multi-objective evolutionary algorithm used in paper \cite{Deb2002AFA}.

In this study, the performance of a counterfactual is represented by a vector of quantitative measures, corresponding to the criteria outlined above. Lower values of the metrics signify better counterfactuals.

For the generation of counterfactuals, the search process plays a critical role. In this study, Gaussian mutation is utilized, with predefined standard deviations assigned to each feature. This process ensures that only a small change will be added to the features each time. The process of generating counterfactual explanations can be summarized into the following steps.

\begin{enumerate}
\item \textbf{Identify target outcome}

Given that the entire road graph contains 3169 road segments, we narrowed its focus to optimizing speed on a single, selected road segment in each experiment. This targeted approach allows for a more manageable and detailed examination of the generated counterfactuals.

\item \textbf{Determine search space}

The search space under consideration is constrained by two key dimensions. The first involves identifying which nodes within the network have features amenable to modification for generating counterfactual explanations. The second aspect focuses on delineating the permissible range within which these counterfactual features can be altered. By establishing these constraints, we create a well-defined scope for generating meaningful and feasible counterfactual explanations.

\item \textbf{Define objective function}

In line with previously outlined criteria, the objective function is constructed as follows:

Let \(f: X\rightarrow\mathbb{R}\) denote the prediction function, \(X^{obs}\) represents the observed feature space, and \(y_{target}\) is the predetermined target speed. A counterfactual explanation \(x'\) for a given observation \(x\)  aims to meet four key criteria: validity, proximity, sparsity, and plausibility. The overarching goal is to minimize a four-component loss function as defined in \cite{molnar2022}:
\begin{equation}
\begin{aligned}
\begin{multlined}
L(x,x',y_{target},X^{obs})=\\
(o_1(f(x'),y_{target}),o_2(x,x'),o_3(x,x'), o_4(x',X^{obs}))
\end{multlined}
\end{aligned}
\end{equation}
where each component captures one of the aforementioned criteria:

\begin{itemize}
    \item \textbf{Validity}: The objective function \( o_1 \) evaluates the distance between the predicted speed \( f(x') \) and the target speed \(y_{target}\):
    \begin{equation}
        o_1(f(x'),y_{target})=|f(x') - y_{target}|
    \end{equation}
    \item \textbf{Proximity}: The objective function \( o_2 \) measures the L1-norm between the original and counterfactual features, \( x \) and \( x' \):
    \begin{equation}
        o_2(x,x')= {\|x - x'\|}_1
    \end{equation}
    \item \textbf{Sparsity}: The objective function \( o_3 \) captures the sparsity of the changes needed to convert \( x \) into \( x' \) by computing the L0-norm:
    \begin{equation}
        o_3(x, x') = \| x - x' \|_0
    \end{equation}
    \item \textbf{Plausibility}: The final objective \( o_4 \) evaluates the plausibility of the counterfactual explanation \( x' \) within the observed feature space \( X^{obs} \). This is calculated by averaging the Euclidean distances between \( x' \) and its \(k\) nearest neighbors in \( X^{obs} \) in an \( n \)-dimensional feature space:
    \begin{equation}
        o_4(x', X^{obs}) = \frac{1}{k} \sum_{i=1}^{k}\sqrt{\sum_{j=1}^{n} (x'_{j}-x^{obs}_{nearest, i, j})^2}
    \end{equation}
    where \(k=3\) in our study.

\end{itemize}

\item \textbf{Searching the Counterfactual Explanations}

The NSGA II is employed to generate a set of counterfactual explanations that satisfy all four objectives. The selection of the most suitable CFE from this set is also a crucial aspect of our approach. 
To facilitate this, an evaluation score \(y_e\) is defined, as shown in Equation \ref{evaluation}. 
This evaluation score serves as a multi-objective trade-off criterion. Users can adjust the weights \( \lambda_1, \lambda_2, \lambda_3, \lambda_4 \) to prioritize specific objectives. For instance, if users value the effectiveness of a CFE in altering the predicted speed over the cost incurred in modifying the features, they might assign a higher weight to the validity objective (\( o_1 \)).
\begin{equation} \label{evaluation}
\begin{multlined}
    y_e= \lambda_1 \frac{o_1}{max(o_1)} + \lambda_2 \frac{o_2}{max(o_2)} + \lambda_3 \frac{o_3}{max(o_3)} + \lambda_4 \frac{o_4}{max(o_4)}
\end{multlined}
\end{equation}

\item \textbf{Evaluating the Counterfactual Explanations}

After the generation and selection of counterfactual explanations, a comprehensive evaluation is essential to understand the generated counterfactuals and assess their performance. 
It is crucial to verify that the counterfactual explanations actually achieve the desired speed improvement for the targeted road segment.
Beyond the targeted road segment, it is also necessary to ensure that localized changes do not negatively impact the speed prediction in other road segments of the network.

\end{enumerate}

\subsection{Scenario-driven counterfactual explanations}

To incorporate user prior constraints effectively, this study proposes an adjustment to the cost function. Specifically, we modified the proximity objective, as represented in Equation \ref{proximity_changed}, to enable the exploration of different scenario settings.

\begin{equation} \label{proximity_changed}
    {o_2'(x,x')} = \sum_{i\neq \bar{E}}^{}\left | x_i - x_i' \right |+\lambda\sum_{i = E}^{}\left | x_i - x_i' \right |
\end{equation}

In this equation, \(E\) represents the feature space that the user wishes to remain unchanged. By incorporating a large weight \( \lambda \), we introduce a significant penalty, steering the generated counterfactual explanations towards user-defined preferences.
This study proposes two distinct mechanisms for integrating user-specific preferences into the counterfactual explanations:

\begin{itemize}
    \item \textbf{Directional Constraints:} Users have the option to specify the direction—either increase or decrease—in which they would like specific features to change. For instance, if the user wants to increase the number of nearby POIs, by setting a large penalty for any generated CFEs where the number of POIs is decreased, the algorithm can tend to generate CFE with a larger number of nearby POIs. 
    
    \item \textbf{Weighting Constraints:} Users can assign weights to individual features to prioritize their importance during the counterfactual generation process. For instance, if a user prefers not to alter the number of lanes on road segments, applying a larger penalty for CFEs where the number of lanes is modified will encourage the algorithm to generate CFEs that maintain the current number of lanes, focusing changes on other features instead.

\end{itemize}

\section{Experiments and results}

The overall performance metrics for the traffic forecasting model are detailed in Table \ref{metricsperformance}.
The \(Accuracy\) reached 91.24 \%, indicating a decent prediction performance. 

\begin{table}[htb]
    \centering
    \begin{tabular}{c|c|c|c|c|c}
         \toprule
   Metrics& RMSE&MAE&Accuracy&\(R^2\)&VAR \\
   \midrule
   Performance&5.7473&2.9876&91.24\% &0.9282&0.9291\\
   
   \bottomrule
    \end{tabular}
    \caption{traffic forecasting model performance.}
    \label{metricsperformance}
\end{table}

\subsection{Generating counterfactual explanations} \label{baseline_exp}
\begin{figure}[htb]
    \centering
    \includegraphics[width=0.8\textwidth]{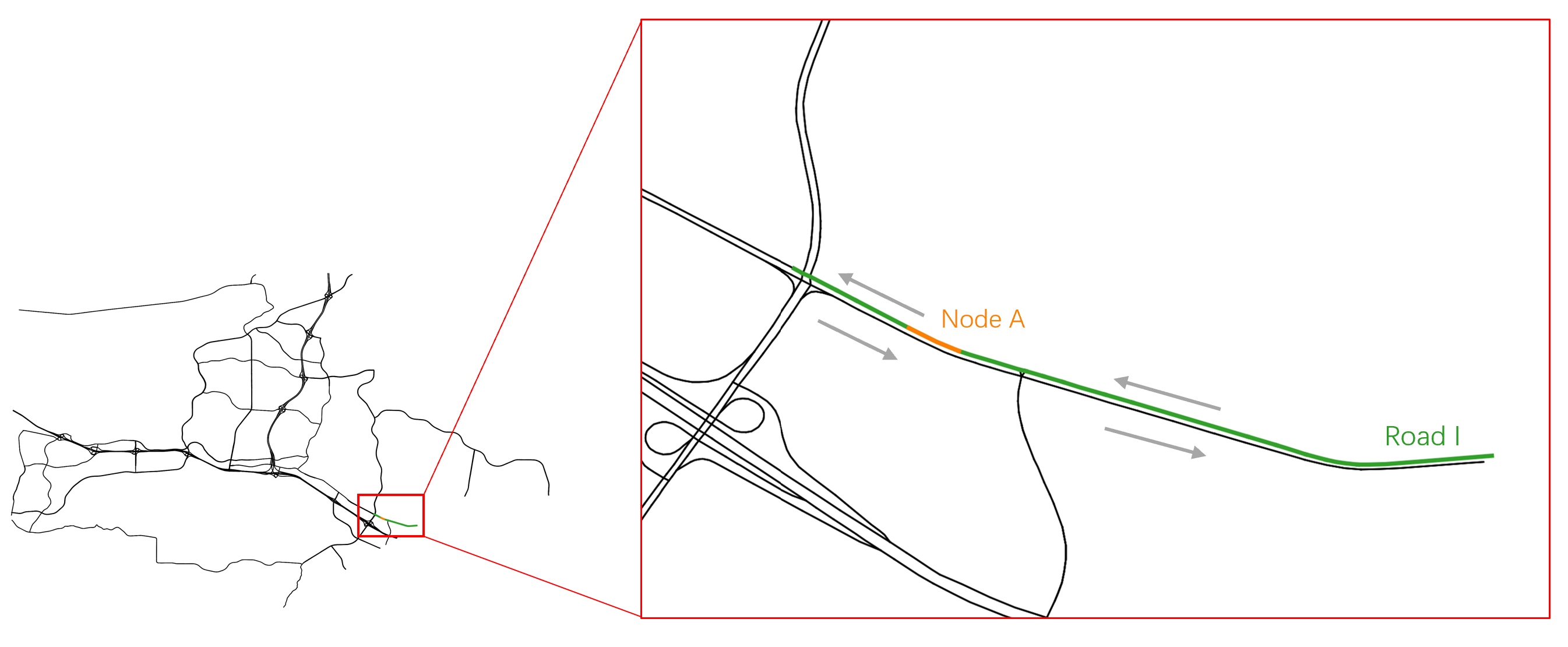}
    \caption{Location of \textit{Node A} and Road I (a suburban road).}
    \label{fig:nodeA}
\end{figure}

\begin{figure}[htb]
    \centering
    \includegraphics[width=0.5\textwidth]{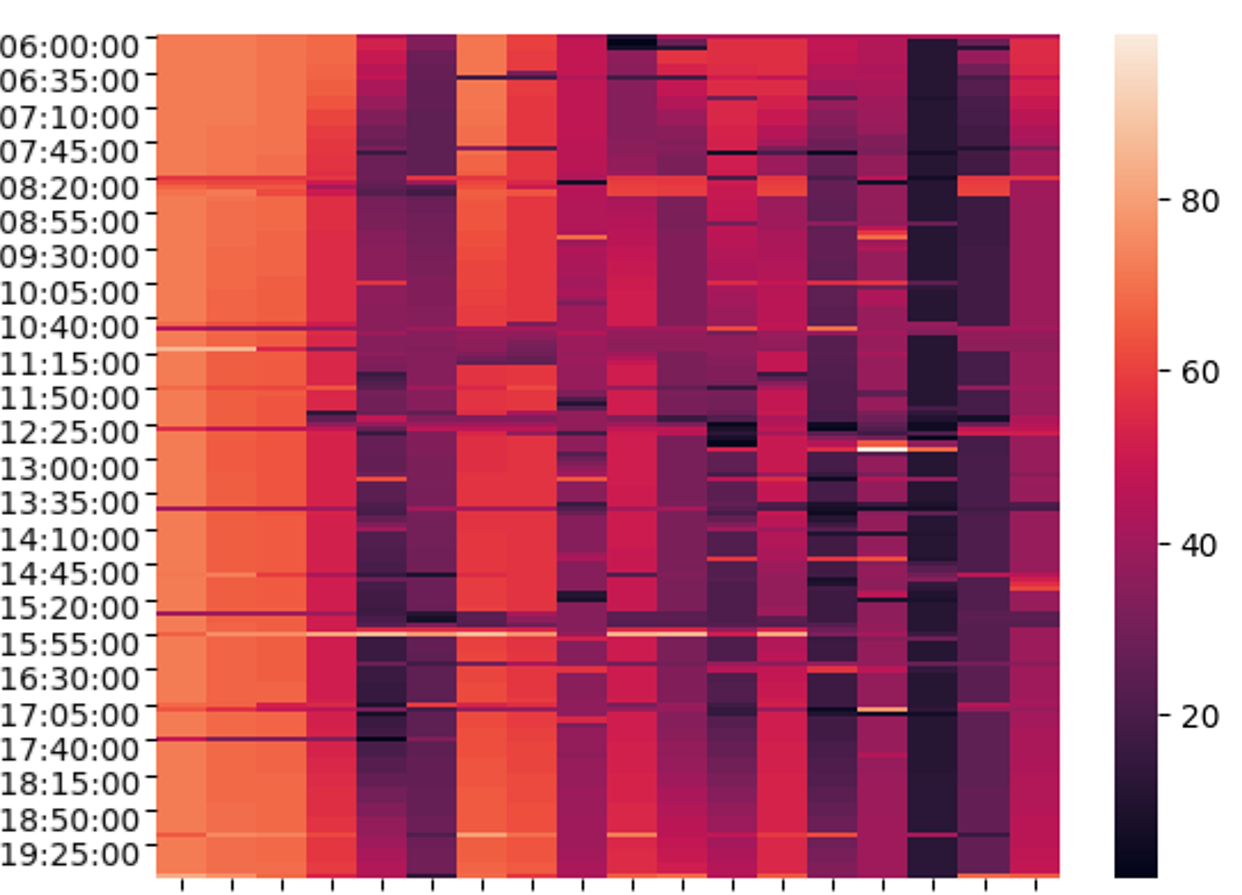}
    \caption{Speed variation for each road segment on Road I on January 10, 2019. The color bar indicates the speed (km/h), each column shows the speed for one road segment, and the traffic flow is from the right side to the left side.}
    \label{fig:RoadIspeed}
\end{figure}

Figure \ref{fig:nodeA} displays the locations of \textit{Node A} on a suburban road, \textit{Road I}, which are the focusing road segments in this experiment. Figure \ref{fig:RoadIspeed} illustrates the speed of each road segment on \textit{Road I} from 6:00 to 8:00, January 10th, 2019.

Specifically, the target of this experiment is to increase the average predicted speed for the road segment \textit{Node A} from 28 km/h to 56 km/h. The prediction uses input data from 8:00 to 8:55 on January 10th, 2019 to predict the traffic speed from 9:00 to 9:55.
Modifications are restricted to road segments situated within \textit{Road I}. 
In this experiment, only the static features of each road segment are considered for modification. 
Based on feature values present in the dataset, the specific ranges for the changeable features are set as follows:
\begin{itemize}
    \item Number of POIs: Range from 0 to 36.
    \item Number of Lanes: Range from 1 to 6.
    \item Speed Limit: Range from 40 to 120 km/h.
\end{itemize}
It is important to note that the speed limit is constrained to remain the same across all segments within \textit{Road I} to be more realistic.

\subsubsection{Objective distributions and correlations}

\begin{figure}[htb]
    \centering 
    \begin{subfigure}{0.3\textwidth}
    \includegraphics[width=\linewidth]{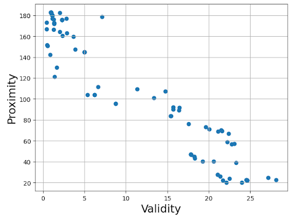}
    \caption{}
    \label{fig:1}
    \end{subfigure}\hfil 
\begin{subfigure}{0.3\textwidth}
  \includegraphics[width=\linewidth]{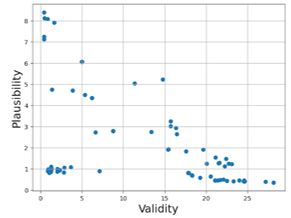}
  \caption{}
  \label{fig:3}
\end{subfigure}\hfil
\begin{subfigure}{0.3\textwidth}
  \includegraphics[width=\linewidth]{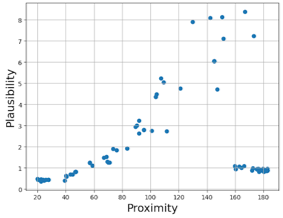}
  \caption{}
  \label{fig:5}  
\end{subfigure}\hfil 

\caption{Objective distribution for the group of counterfactual explanations. (a) shows the distribution between validity and proximity; (b) shows the distribution between validity and plausibility; (c) shows the distribution between proximity and plausibility.}
\label{fig:objectives}
\end{figure}

Figure \ref{fig:objectives} shows the distribution of the objectives for the set of counterfactual explanations generated in this experiment. The distribution patterns reveal insights into the relationships among different objectives.

\paragraph{Validity - Proximity}
As illustrated in Figure \ref{fig:1}, there appears to be a negative correlation between the validity loss and the proximity loss, which suggests that as counterfactual predictions become closer to the target speed, the divergence of the generated counterfactual features from the original features increases. 

\paragraph{Validity - Plausibility}
Similar observations can be made from Figure \ref{fig:3}, where validity loss and plausibility loss are negatively correlated. This implies that when the counterfactual predictions become closer to the target speed, they tend to deviate more from observed points in the feature space. 

\paragraph{Proximity - Plausibility}
Figure \ref{fig:5} depicts an overall positive correlation between proximity loss and plausibility loss. Generally, a greater proximity loss is accompanied by a larger plausibility loss. However, an interesting cluster of points exists in the bottom-right corner of this figure. These points show that there are counterfactual explanations that differ substantially from the original features but still maintain an overall close distance to observed data points.

\subsubsection{Evaluation of the most optimal counterfactual explanations}
Different weight parameters can be assigned to each objective function in Equation \ref{evaluation} to find the optimal counterfactual explanation for a particular interest or purpose. As a case study, we investigated the results where \(\lambda_1 = 1\), \(\lambda_2 = 0.2\), \(\lambda_3 = 0.2\), and \(\lambda_4 = 0.6\). This choice of weights reflects the relative importance of different criteria in the evaluation score. 
Particularly, validity is prioritized as the most critical factor and is assigned the highest weight. Plausibility also holds significance, but to a lesser extent, so it was assigned a weight of 0.6. Given that sparsity was considered less crucial for this particular study, it was given a lower weight of 0.2. Additionally, since proximity and plausibility are interrelated, we assigned proximity a smaller weight of 0.2 to ensure a balanced evaluation.

The optimal counterfactual explanation with the given weights produces the objective scores outlined in Table \ref{objectivesore}. The validity score shows a minimal deviation of 1.3496 km/h from the target speed. Regarding sparsity, a total of 32 features were altered across all road segments on \textit{Road I}. 

\begin{table}[htb]

    \centering
    \begin{tabular}{c|c|c|c}
         \toprule
   Validity (\(o_1\))&Proximity (\(o_2\))&Sparsity (\(o_3\))&Plausibility (\(o_4\)) \\
   \midrule
   1.3496 &	172.5508&	32&	 0.9862\\
 
   \bottomrule
    \end{tabular}
    \caption{Objective value for the selected counterfactual explanation.}
    \label{objectivesore}
\end{table}

Table \ref{speedtarget} shows the speed prediction change with this optimal counterfactual explanation. With the counterfactual features, the speed prediction increases significantly and is very close to the target speed of 56 km/h.

\begin{table}[htb]
    \centering
    \begin{tabular}{c|c|c}
         \toprule
   Original prediction & Counterfactual prediction & Target\\
   \midrule
   30.10 km/h &	54.65 km/h &	56 km/h\\
 
   \bottomrule
    \end{tabular}
    \caption{Average speed prediction from 9:00 to 10:00, January 10th, 2019.}
    \label{speedtarget}
\end{table}

\begin{figure}[htb]
     \centering
     \begin{subfigure}[htb]{0.4\textwidth}
         \centering
         \includegraphics[width=\textwidth]{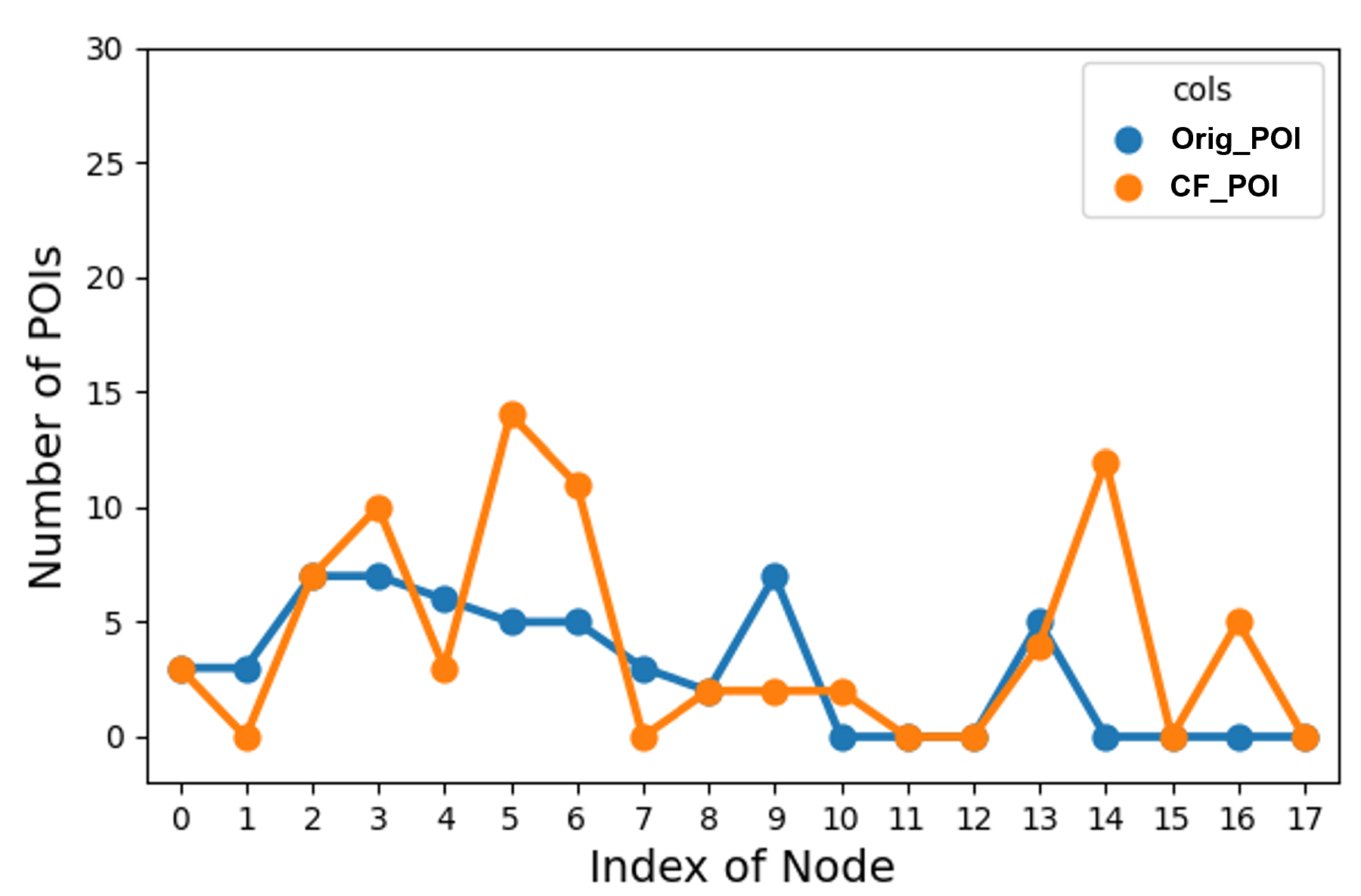}
         \caption{Comparing number of POIs.}
         \label{fig:cf_poi}
     \end{subfigure}
     \begin{subfigure}[htb]{0.4\textwidth}
         \centering
         \includegraphics[width=\textwidth]{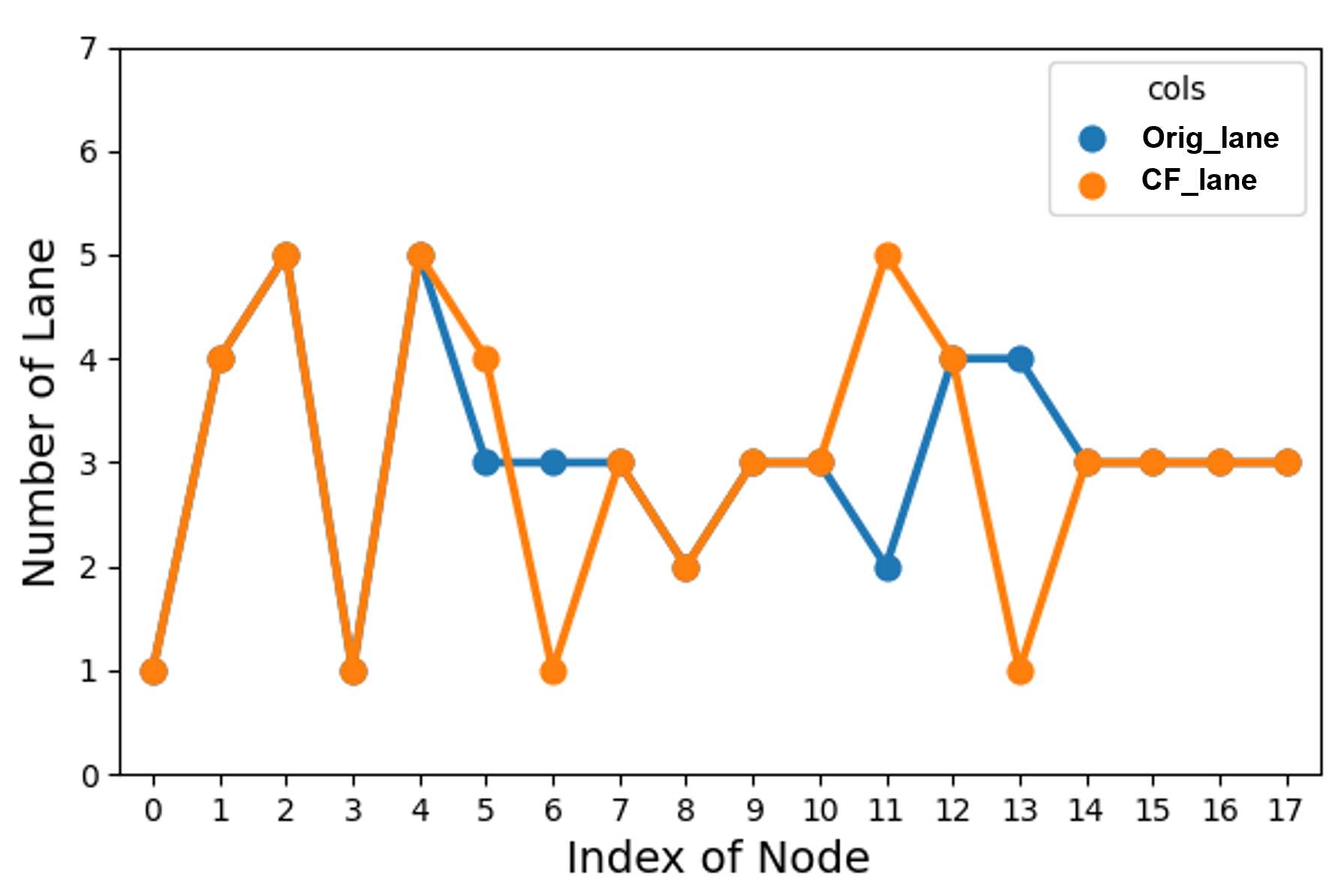}
         \caption{Comparing number of lanes.}
         \label{fig:cf_lane}
     \end{subfigure}
     \caption{Comparison between original features (in blue) and counterfactual features (in orange) for each road segment on \textit{Road I}. The x-axis represents individual road segments and is arranged to follow the direction of traffic flow.}
     \label{fig:cf_nodeA_example}
\end{figure}

Figure \ref{fig:cf_nodeA_example} shows the comparison between original features and the selected counterfactual features (the number of POIs and the number of lanes) for each road segment on \textit{Road I}. The counterfactual features in Figure \ref{fig:cf_poi} suggest that a general increase in POIs at certain locations of the road network is associated with higher speed prediction. Given other counterfactual features, the number of lanes only needs minor modification at a few locations to achieve the target speed (Figure \ref{fig:cf_lane}). The original speed limit is 72 km/h, while the counterfactual speed limit is 105.62 km/h. 

\subsection{Spatial comparison}

\begin{figure}[h!]
    \centering
    \includegraphics[width=0.8\textwidth]{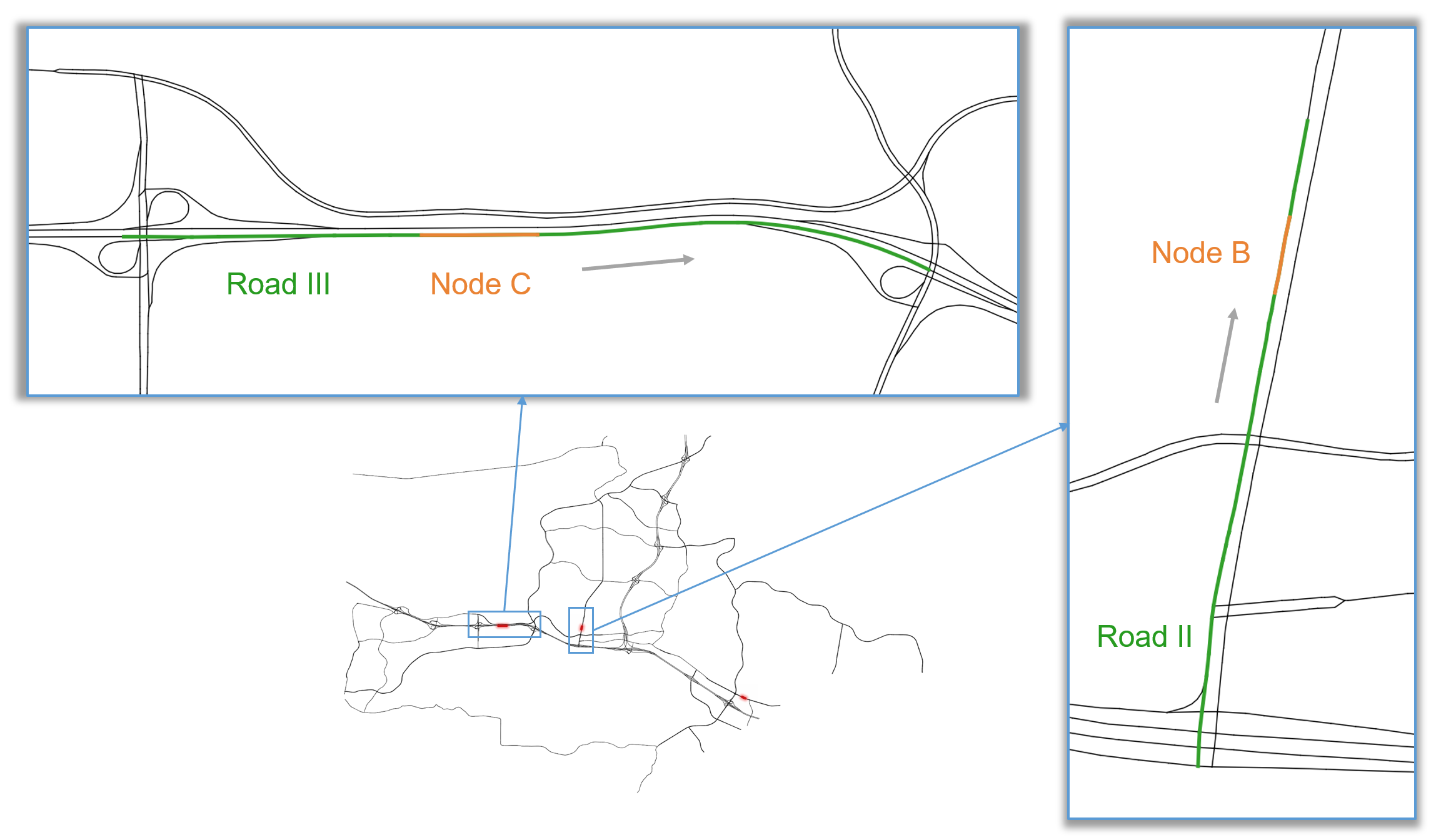}
    \caption{Location of \textit{Node B} and \textit{Node C}. \textit{Node B} is located on an urban road, \textit{Node C} is located on a highway.}
    \label{fig:locationBC}
\end{figure}

\begin{figure}[h!]
    \centering
    \includegraphics[width=0.75\textwidth]{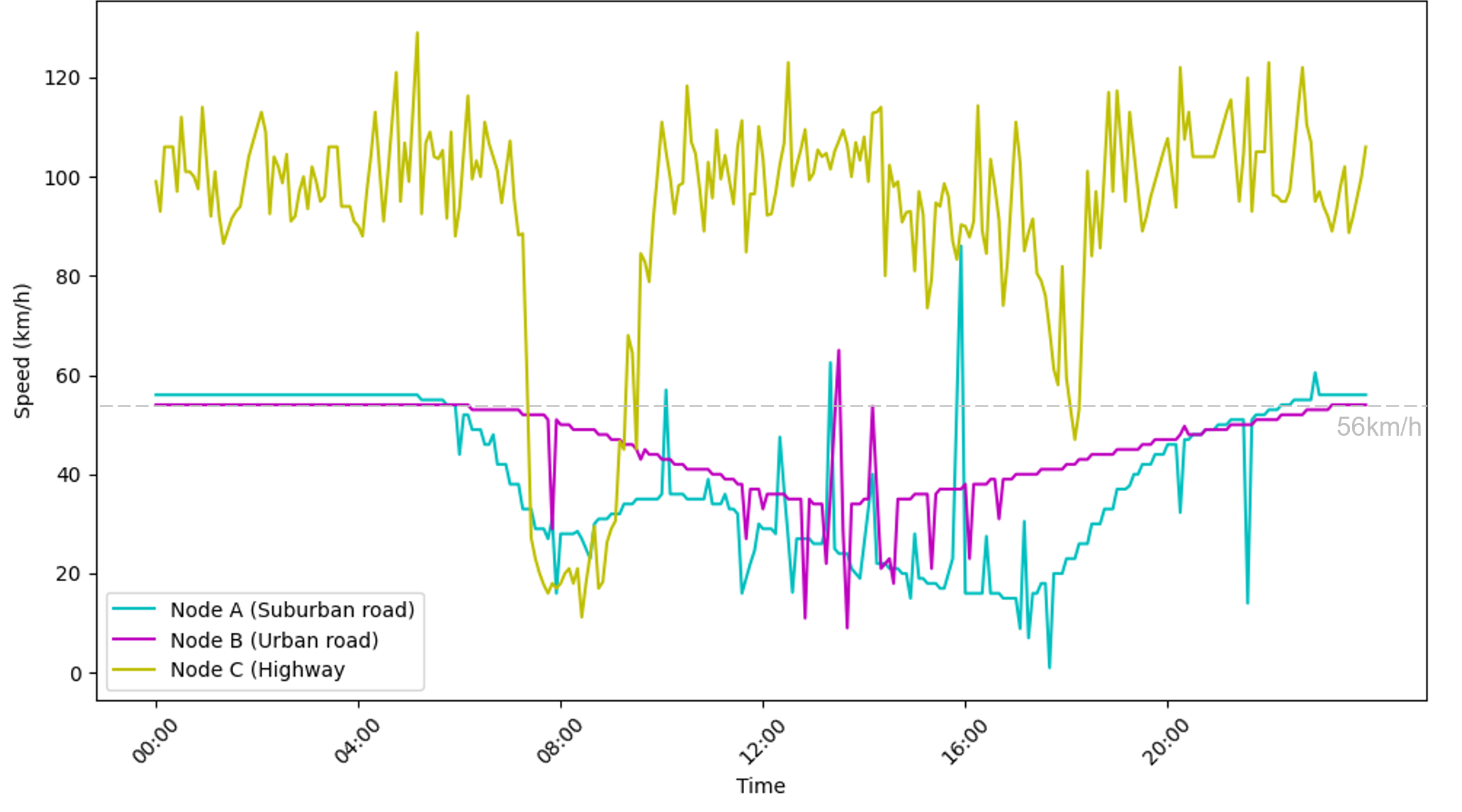}
    \caption{Speed of \textit{Node A}, \textit{Node B}, and \textit{Node C} on January 10th, 2019. The gray dashed line indicates the target speed of 56 km/h for generating counterfactuals. }
    \label{fig:speed_nodeABC}
\end{figure}

The type of road facility (e.g., highway, urban road, or suburban road) is widely acknowledged as an important factor influencing traffic patterns \cite{inproceedingsroadtype}. In light of this, to gain deeper insights into how the deep learning model predicts speed differently across different types of roads, this section compares the counterfactual explanations generated for three distinct types of road segments, i.e., a suburban road, an urban road, and a highway, represented by \textit{Node A}, \textit{Node B}, and \textit{Node C} respectively. Figure \ref{fig:locationBC} displays the locations of the two additional nodes, \textit{Node B} and \textit{Node C}. Figure \ref{fig:speed_nodeABC} shows the speeds of the three nodes on January 10th, 2019.

\begin{figure}[h!]
     \centering
     \begin{subfigure}{0.3\textwidth}
         \centering
         \includegraphics[width=\textwidth]{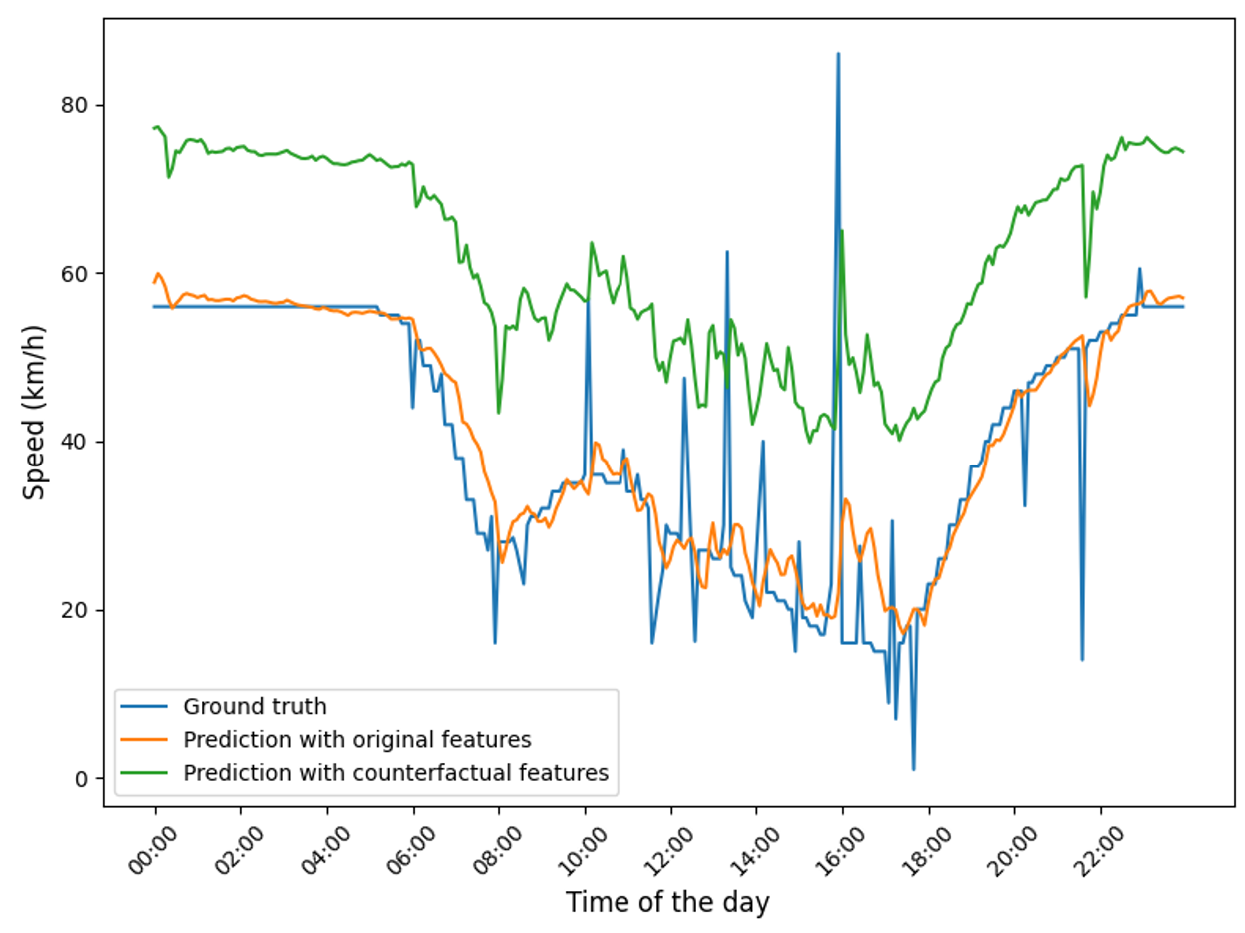}
         \caption{Node A}
         \label{fig:nodeA_day}
     \end{subfigure}
     \hfill
     \begin{subfigure}{0.3\textwidth}
         \centering
         \includegraphics[width=\textwidth]{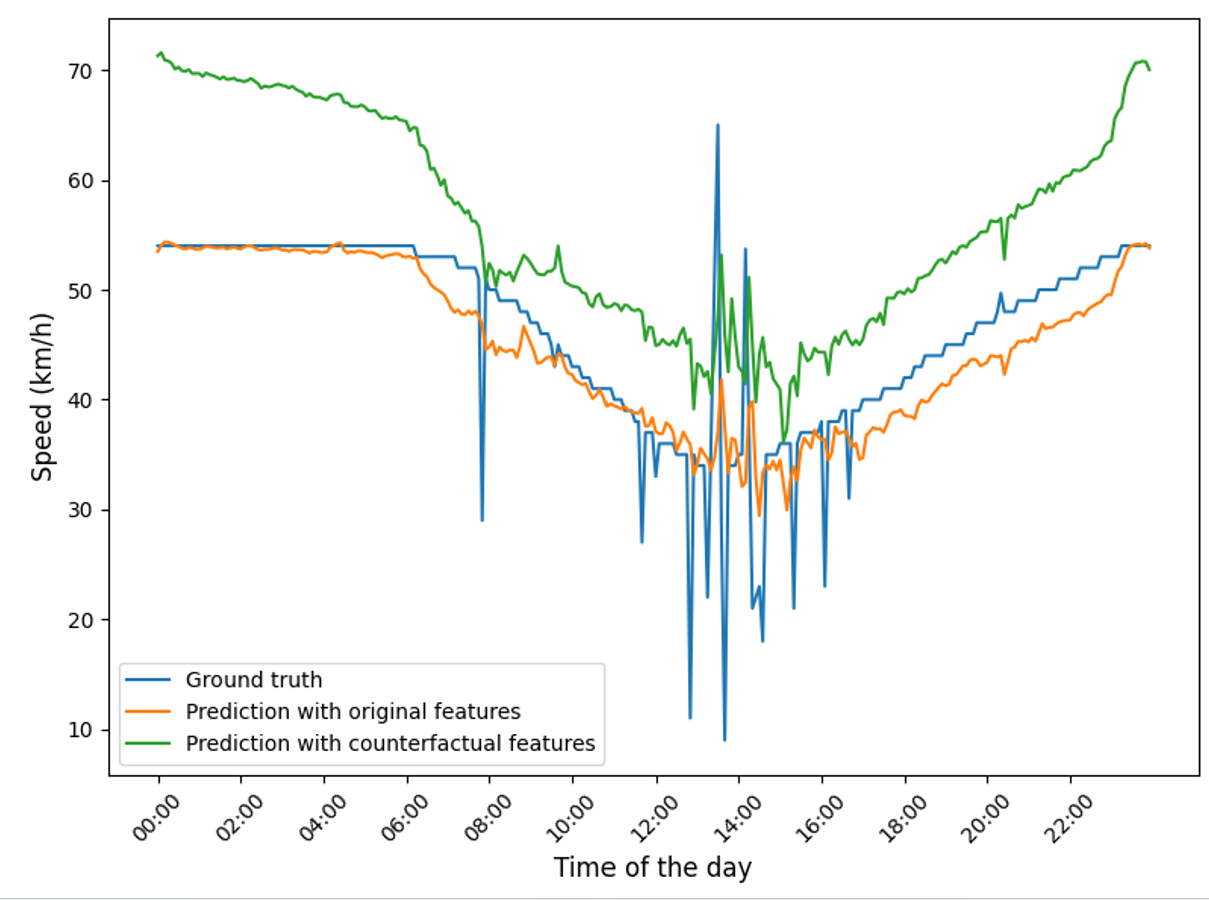}
         \caption{Node B}
         \label{fig:nodeB_day}
     \end{subfigure}
     \hfill
     \begin{subfigure}{0.3\textwidth}
         \centering
         \includegraphics[width=\textwidth]{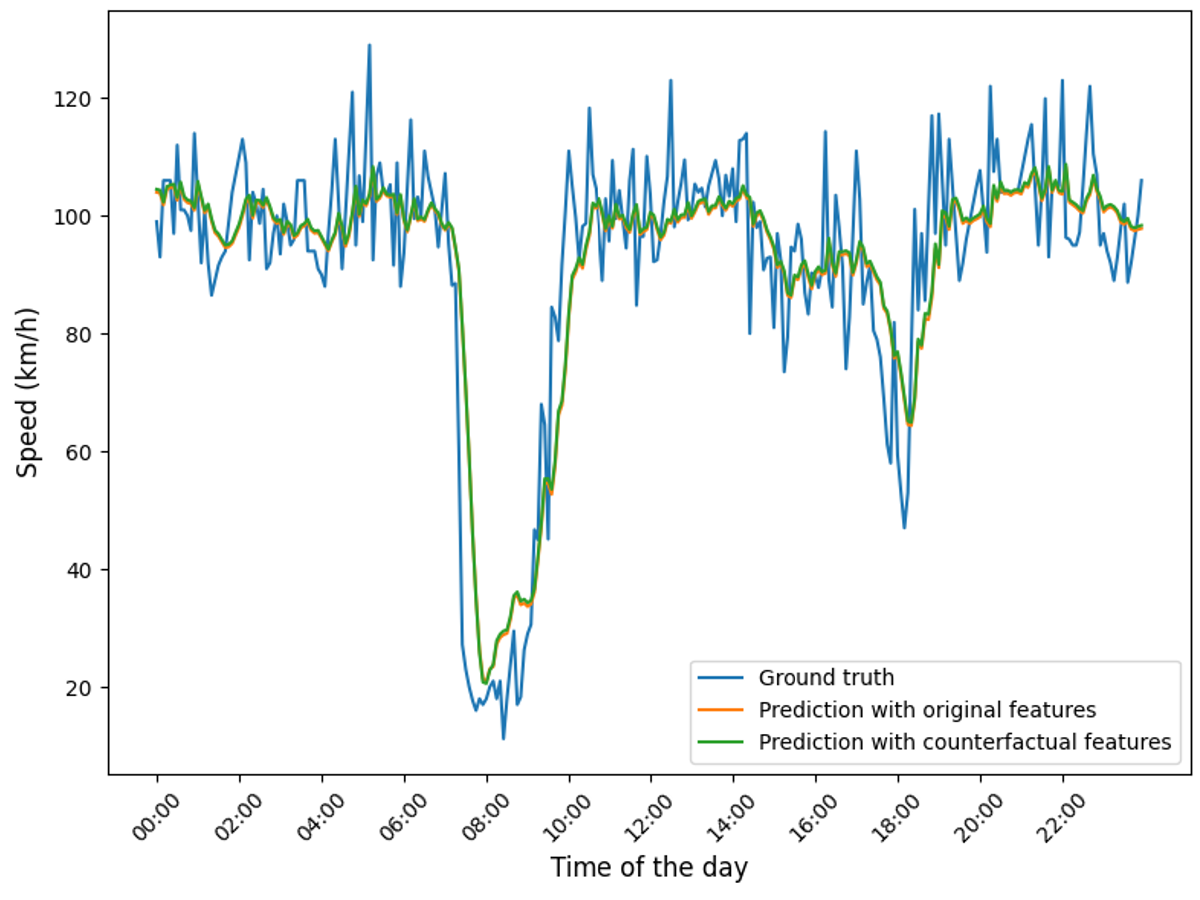}
         \caption{Node C}
         \label{fig:nodeC_day}
     \end{subfigure}
        \caption{Comparison of ground truth speed, original speed prediction, and counterfactual speed prediction for \textit{Node A}-suburban road, \textit{Node B}-urban road, \textit{Node C}-highway on January 10, 2019. }
        \label{fig:spatialcompareABC}
\end{figure}

For all three road segments, the target was set identically: to increase the predicted average speed on each node between 9:00 and 10:00 to 56 km/h. The initial average speeds recorded were 28.2 km/h for Node A, 49 km/h for Node B, and 20.18 km/h for Node C. To achieve the target speed, counterfactual explanations were generated and selected for each node following the procedures outlined in Section \ref{MOC_objective}. To compare the impact of the generated counterfactual features on the daily pattern, we generate counterfactual predictions for each node for the entire day and display the results respectively in Figure \ref{fig:spatialcompareABC}. 

Figure \ref{fig:nodeA_day} and Figure \ref{fig:nodeB_day} reveal that the generated counterfactual explanations for node A (suburban road) and node B (urban road) managed to increase the predicted speed, particularly for the targeted duration (9:00 - 10:00). However, Figure \ref{fig:nodeC_day} shows that the counterfactual explanation for node C (highway) did not result in a substantial speed increase. This demonstrates that static features, including the number of POIs, the number of lanes, and speed limits, do not exert a significant influence on predicting highway speeds. Therefore, in the following experiments, we will only focus on \textit{Node A} and \textit{Node B} for subsequent analyses.

\subsubsection{Evaluating the global impact of counterfactuals on the traffic network}

Since we only generated counterfactual features at local road segments to increase the predicted speed on a particular node, it is uncertain whether the generated counterfactuals will negatively impact predicted traffic in other parts of the road network. In this section, we evaluate the global impact of counterfactual explanations on the speed prediction for the entire traffic network.

\begin{figure}[h!]
     \centering
     \begin{subfigure}{0.6\textwidth}
         \centering
         \includegraphics[width=\textwidth]{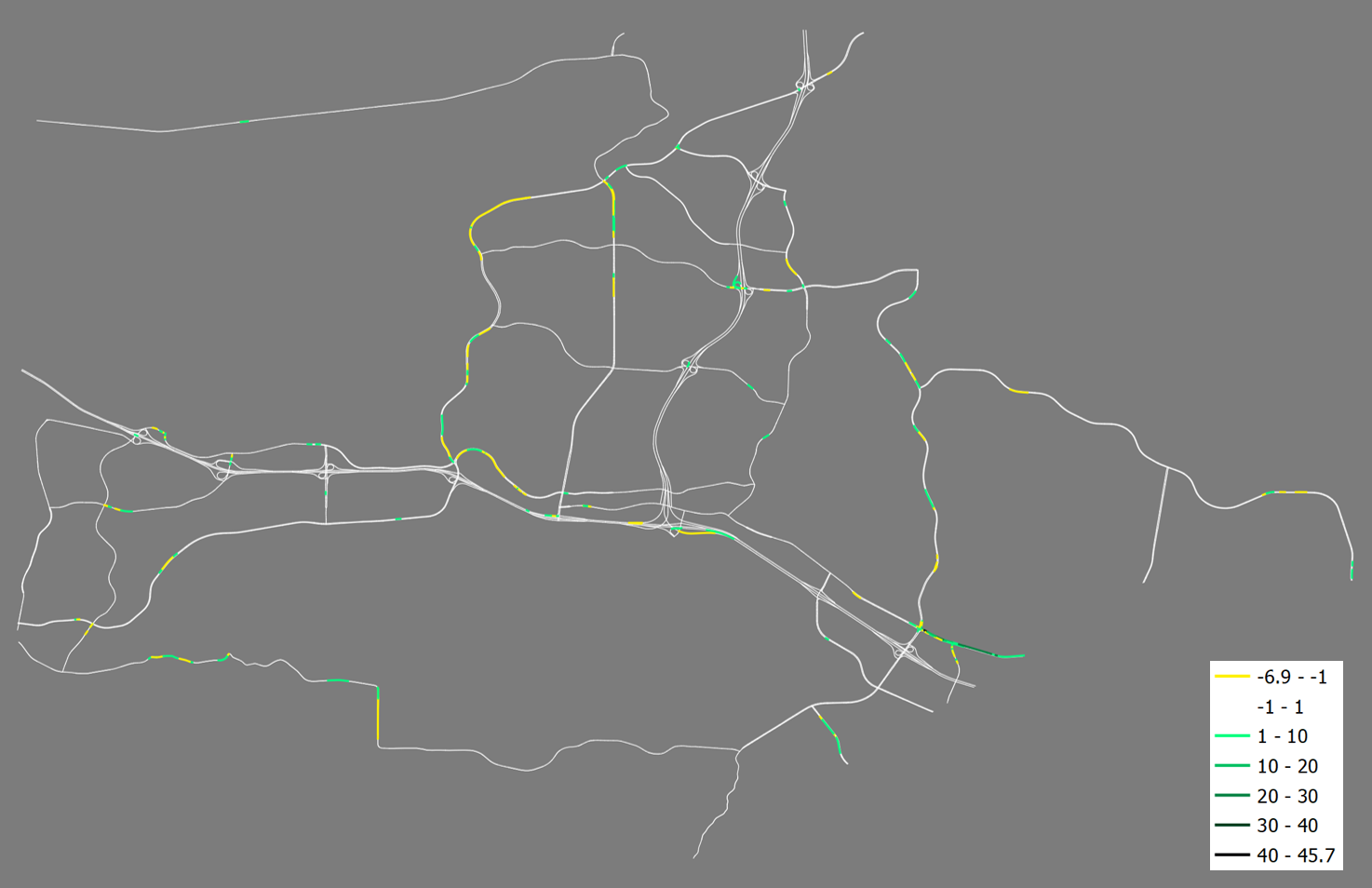}
    \caption{Impact of CFE for \textit{Node A}.}
         \label{fig:allgraphimpactA}
     \end{subfigure}
     \begin{subfigure}{0.6\textwidth}
         \centering
         \includegraphics[width=\textwidth]{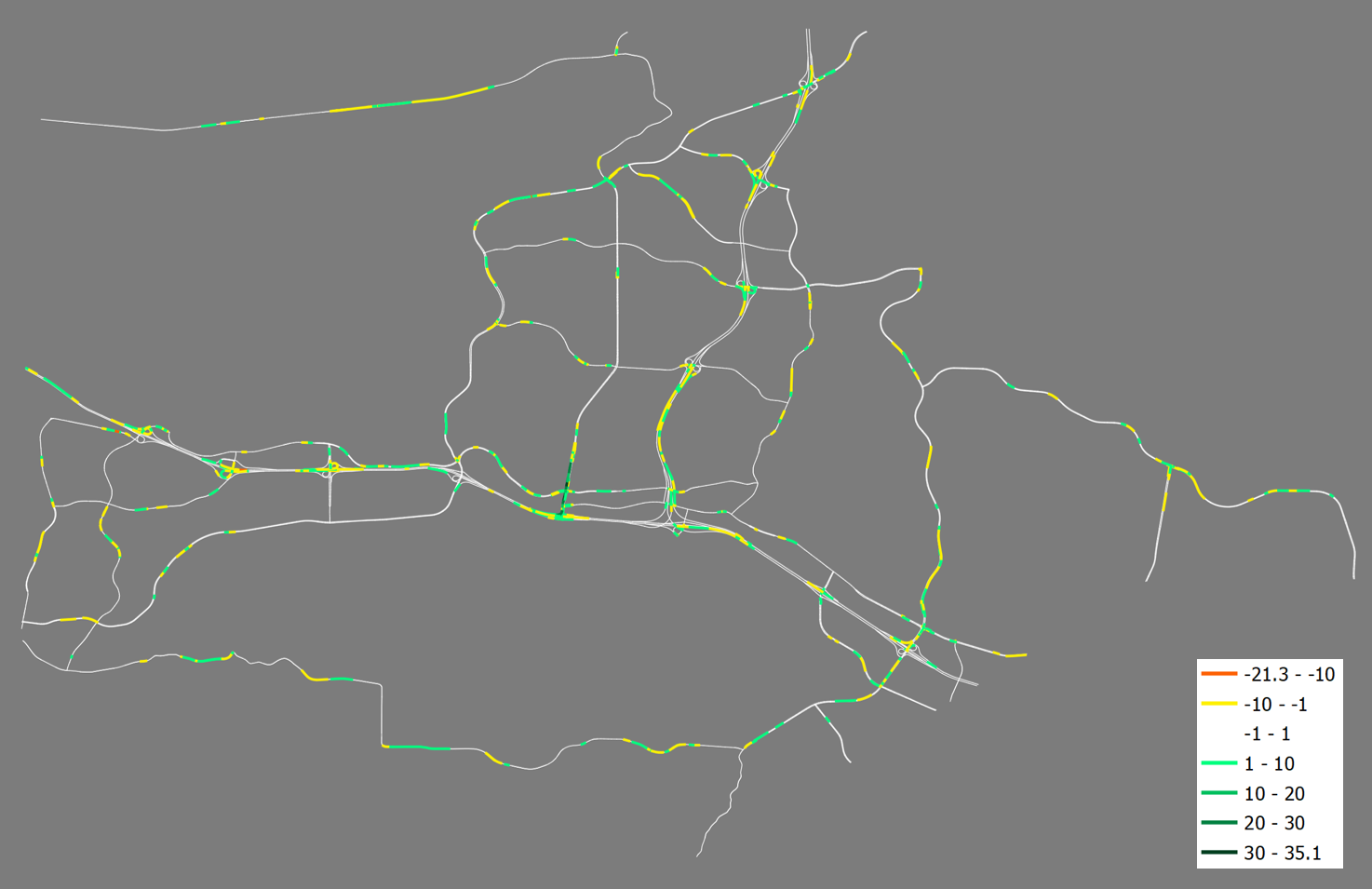}
    \caption{Impact of CFE for \textit{Node B}.}
         \label{fig:allgraphimpactB}
     \end{subfigure}
     \caption{Difference between original speed prediction and counterfactual speed prediction (km/h).}
     \label{fig:allgraphimpact}
\end{figure}

Figure \ref{fig:allgraphimpact} shows the difference between the counterfactual speed prediction and the original speed prediction. 
In Figure \ref{fig:allgraphimpactA} the speed increase is mainly distributed on the urban road \textit{Road I}. The counterfactual features only have a minimal negative impact on the speed prediction of other locations, with a maximum decrease of 6.9 km/h in predicted speed. 
In contrast, Figure \ref{fig:allgraphimpactB} shows that the counterfactuals generated for \textit{Node B} on the urban road also broadly change the predicted traffic speed in other road segments. In addition, the negative impact caused by counterfactuals at \textit{Node B} (urban road) is larger than those at \textit{Node A} (suburban road). The largest speed decrease reaches 21.3 km/h with the counterfactual features generated for urban roads.

\subsection{Temporal comparison}

Temporal setting also significantly influences traffic patterns. To examine these effects, we compared counterfactuals generated for five time slots:

\begin{itemize}
    \item \textbf{Morning} Jan 10th (Thursday): 8:00 – 10:00
    \item \textbf{Noon} Jan 10th (Thursday): 12:00 – 14:00
    \item \textbf{Afternoon} Jan 10th (Thursday): 15:00 – 17:00
    \item \textbf{Evening} Jan 10th (Thursday): 18:00 – 20:00
    \item \textbf{Weekend} Jan 13th (Sunday): 8:00 – 10:00
\end{itemize}

\begin{figure}[htb]
     \centering
     \begin{subfigure}{0.9\textwidth}
         \centering
         \includegraphics[width=\textwidth]{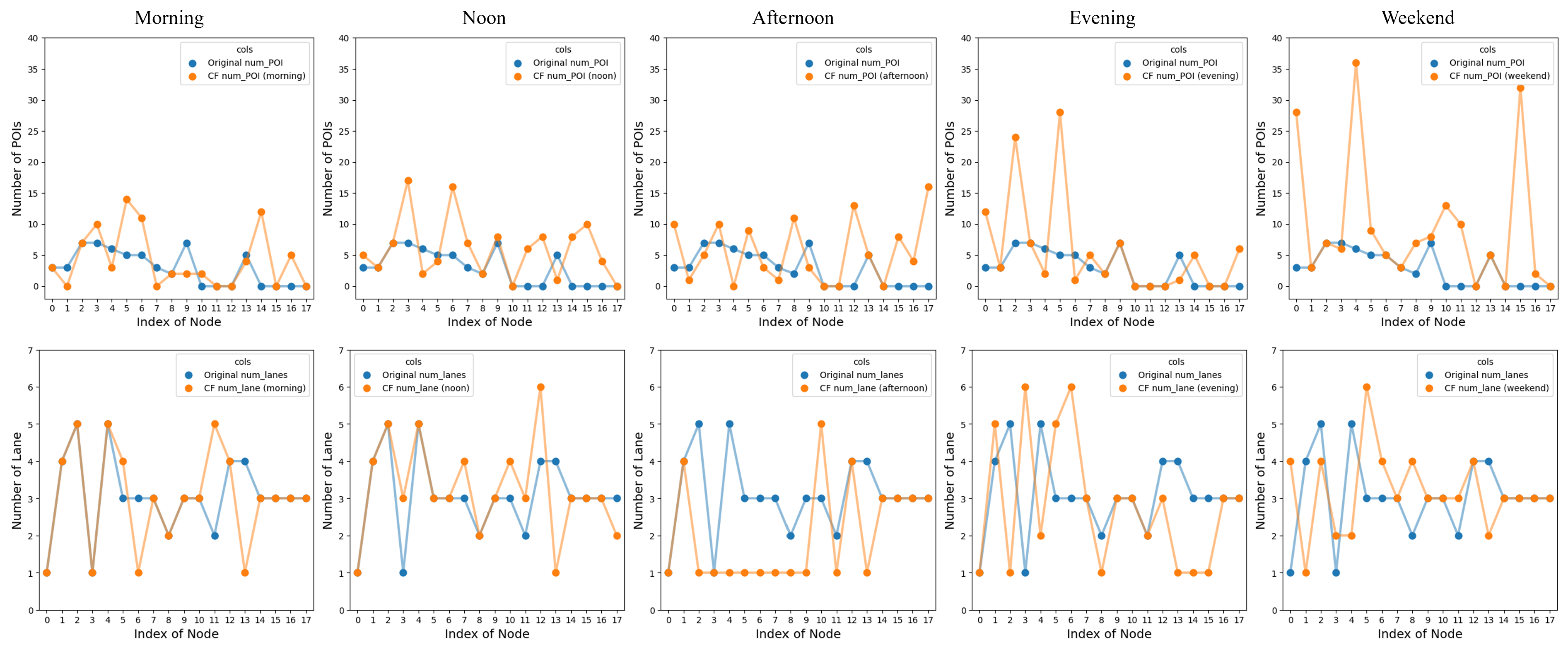}
         \caption{Comparing temporal settings on \textit{Node A}.}
         \label{fig:nodea_temporal}
     \end{subfigure}
     \begin{subfigure}{0.9\textwidth}
         \centering
         \includegraphics[width=\textwidth]{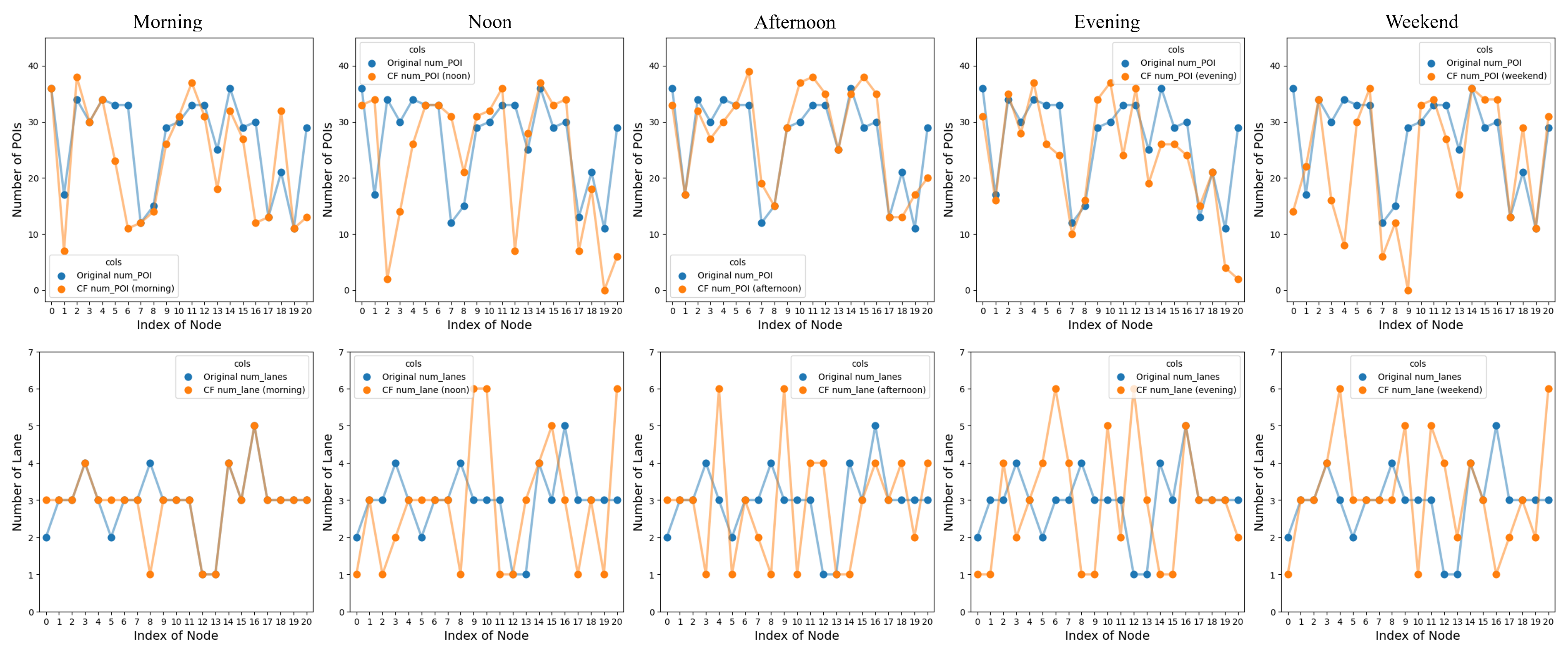}
         \caption{Comparing temporal settings on \textit{Node B}.}
         \label{fig:nodeb_temporal}
     \end{subfigure}
     \caption{Comparison between original and counterfactual number of POIs and number of lanes for different temporal settings on \textit{Node A} and \textit{Node B}.}
     \label{fig:nodeAB_temporal_cf}
\end{figure}

\begin{figure}[htb]
     \centering
     \begin{subfigure}{0.3\textwidth}
         \centering
         \includegraphics[width=\textwidth]{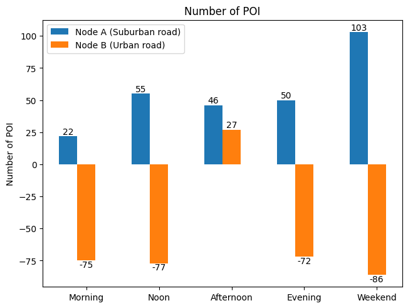}
         \caption{}
         \label{fig:poi_temporal}
     \end{subfigure}
     \hfill
     \begin{subfigure}{0.3\textwidth}
         \centering
         \includegraphics[width=\textwidth]{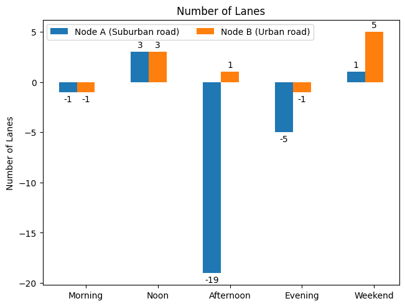}
         \caption{}
         \label{fig:lane_temporal}
     \end{subfigure}
     \hfill
     \begin{subfigure}{0.3\textwidth}
         \centering
         \includegraphics[width=\textwidth]{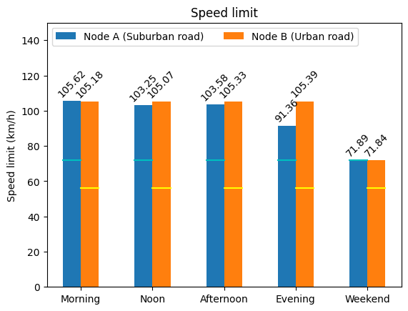}
         \caption{}
         \label{fig:spdlimit_temporal_compare}
     \end{subfigure}
     \caption{Comparison of the difference between counterfactual and original features in different temporal settings for \textit{Node A} and \textit{Node B}. (a) shows the total difference between the counterfactual and the original number of POIs; (b) shows the total difference between the counterfactual and the original number of lanes; (c) shows the counterfactual speed limit (the original speed limit on \textit{Node A} is 72 km/h, the original speed limit on \textit{Node B} is 56 km/h).}
     \label{fig:nodeAB_temporal_cf_stat}
\end{figure}

We generated counterfactual explanations for \textit{Node A} on the suburban road and \textit{Node B} on the urban road during each of these time slots. The most optimal counterfactual explanations for each temporal setting across all nodes in the road segment are illustrated in Figure \ref{fig:nodeAB_temporal_cf}. Summing over the difference across all nodes, Figure \ref{fig:nodeAB_temporal_cf_stat} compares the total difference between the counterfactual features and original features for each setting.

\subsubsection{Comparison of number of POIs}

Figure \ref{fig:poi_temporal} illustrates the variations in the counterfactual number of POIs for both \textit{Node A} and \textit{Node B} across the selected time slots.

\paragraph{Node A:}
The counterfactual features for \textit{Node A} show a consistent increase in the number of POIs across all time slots. This trend suggests that the model associates a higher number of POIs with lower congestion levels on suburban roads. Interestingly, this increase is more pronounced during weekends, indicating that during weekends, the number of POIs has a stronger influence on the speed of suburban roads.

\paragraph{Node B:}
On the other hand, for \textit{Node B}, which is located on an urban road, Figure \ref{fig:nodeAB_temporal_cf_stat} reveals that the counterfactual explanations generally advocate for a reduction in the number of POIs. This can be attributed to the high original count of nearby POIs, which likely contribute to traffic congestion. Thus, reducing the number of POIs is suggested to mitigate traffic demand. However, it is worth noting that, in the afternoon setting, the counterfactual number of POIs stays relatively consistent, which can be interpreted that in the weekday afternoon, the number of POIs has a small impact on the traffic of urban roads.

\subsubsection{Comparison of number of lanes}

If we compare the difference between the counterfactual number of lanes and the original number of lanes in Figure \ref{fig:lane_temporal}. There are no substantial changes for most time slots in both \textit{Node A} and \textit{Node B}. However, in the afternoon on the suburban road, the number of lanes drops by 19 compared to the original number of lanes, with most reduction occurring in the upstream part of the road segment. This can be interpreted that in the afternoon on the suburban road, counterfactual explanations suggest a decrease in the number of lanes (node index 0 to 10) before node B (index 11), thereby limiting the volume of cars and enabling smoother traffic flow.

\subsubsection{Comparison of speed limit}
Figure \ref{fig:spdlimit_temporal_compare} presents counterfactual speed limits for each setting. For \textit{Node A}, the speed limit increases for all time slots except on the weekend, implying that changing the speed limit may not be effective on suburban roads during this period. For \textit{Node B}, the counterfactual speed limits remain fairly consistent throughout weekdays but drop on weekends, possibly due to lower congestion levels. 

\subsection{Experiments on scenario-driven counterfactual explanations}
\subsubsection{Directional constraints}

Directional constraints allow users to specify the desired direction of feature change—either an increase or a decrease. In the scope of this experiment, several scenario-specific constraints are evaluated and compared. We focus on \textit{Node A} on the suburban road for demonstration.
The objective is to enhance the predicted speed between 9:00 and 10:00 to reach 56 km/h.

\begin{figure}[htb]
     \centering
     \begin{subfigure}{0.3\textwidth}
         \centering
         \includegraphics[width=\textwidth]{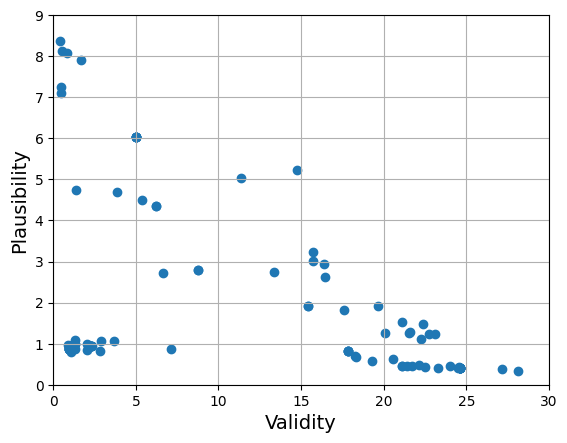}
         \caption{}
         \label{fig:}
     \end{subfigure}
     \hfill
     \begin{subfigure}{0.3\textwidth}
         \centering
         \includegraphics[width=\textwidth]{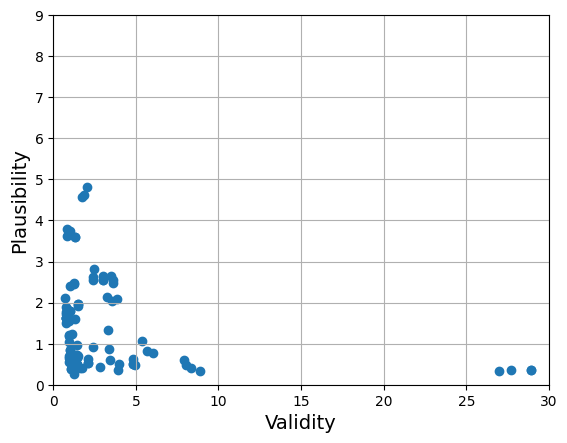}
         \caption{}
         \label{fig:}
     \end{subfigure}
     \hfill
     \begin{subfigure}{0.3\textwidth}
         \centering
         \includegraphics[width=\textwidth]{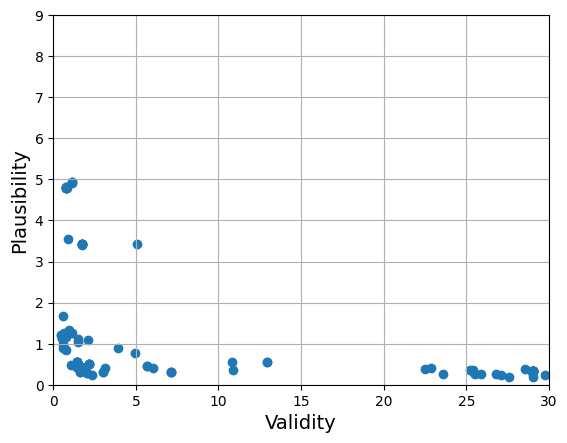}
         \caption{}
         \label{fig:}
     \end{subfigure}
     \caption{Objective distribution (Validity v.s. Plausibility) for different directional constraint settings. (a) has no scenario constraint; (b) has scenario constraints on the number of POIs decreasing and the number of lanes increasing; (c) has scenario constraints on the number of POIs increasing.}
     \label{fig:objectivedirectional}
\end{figure}

Despite the additional requirement on the direction of feature change, we want to ensure that the generated counterfactual explanations achieve the desired prediction (i.e., low validity loss) and are close to the feature space of the observational data (i.e., low plausibility loss). Therefore, we examined the validity and plausibility scores of scenario-based counterfactuals, as shown in Figure \ref{fig:objectivedirectional}. The figure shows even with the directional constraint, the distribution of the two objective scores falls within a similar range as the one without directional constraint. 

\begin{figure}[htb]
  \centering
  \renewcommand{\arraystretch}{1}
  \begin{tabularx}{1.0\linewidth}{@{}
      l
      X @{\hspace{6pt}}
      X @{\hspace{6pt}}
      X
    @{}}
    & \multicolumn{1}{c}{Scatter}
    & \multicolumn{1}{c}{Interpolation} 
    & \multicolumn{1}{c}{3D scatter} \\
    \makebox[20pt]{\raisebox{40pt}{\rotatebox[origin=c]{90}{Scenario A}}}%
    & \includegraphics[width=0.3\textwidth]{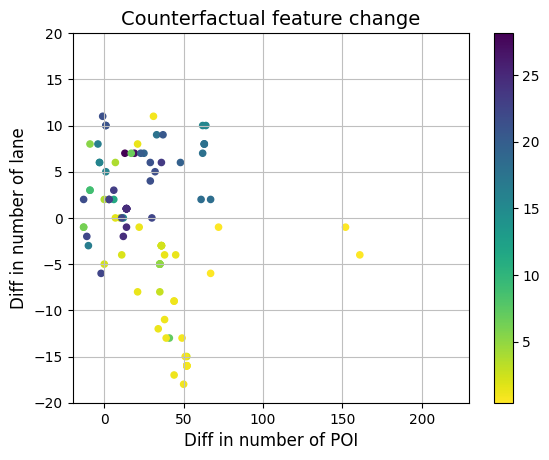}
    & \includegraphics[width=0.3\textwidth]{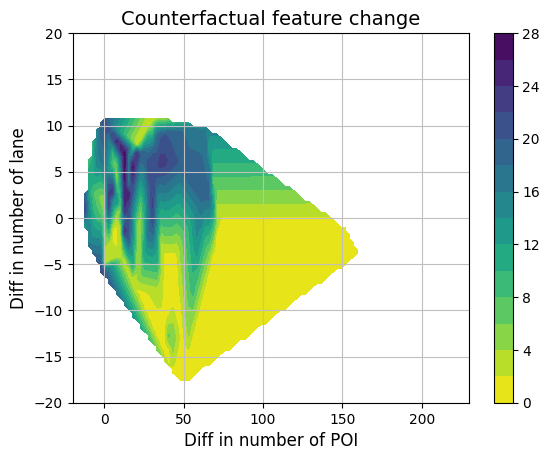} 
    & \includegraphics[width=0.3\textwidth]{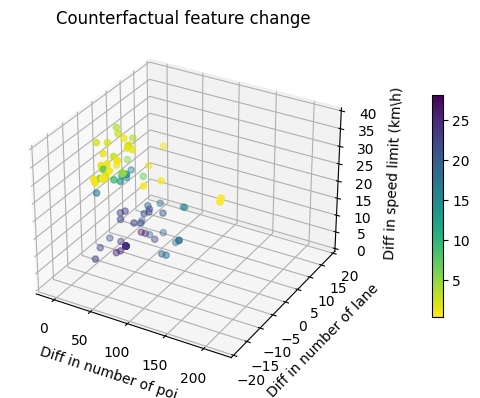}\\
    \makebox[20pt]{\raisebox{40pt}{\rotatebox[origin=c]{90}{Scenario B}}}%
    & \includegraphics[width=0.3\textwidth]{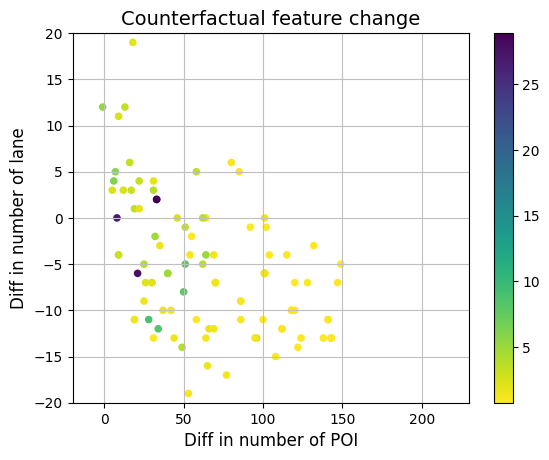}
    & \includegraphics[width=0.3\textwidth]{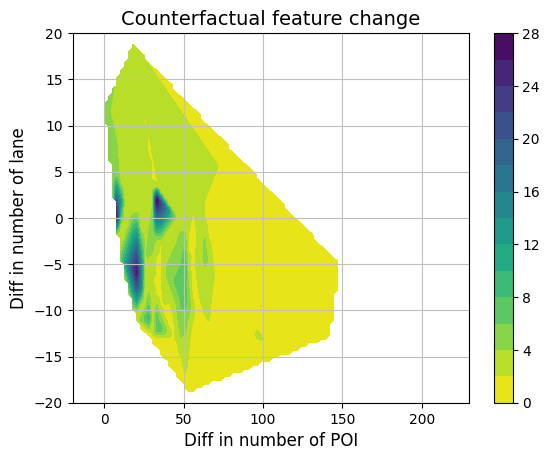}
    & \includegraphics[width=0.3\textwidth]{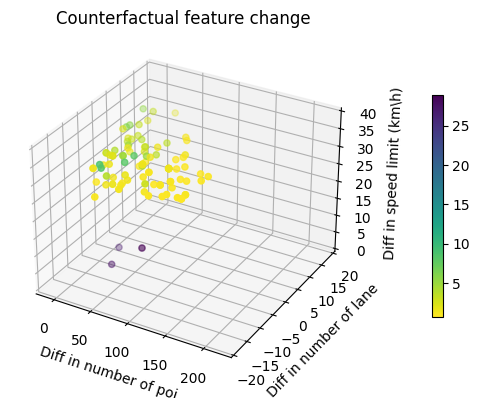} \\
    \makebox[20pt]{\raisebox{40pt}{\rotatebox[origin=c]{90}{Scenario C}}}%
    & \includegraphics[width=0.3\textwidth]{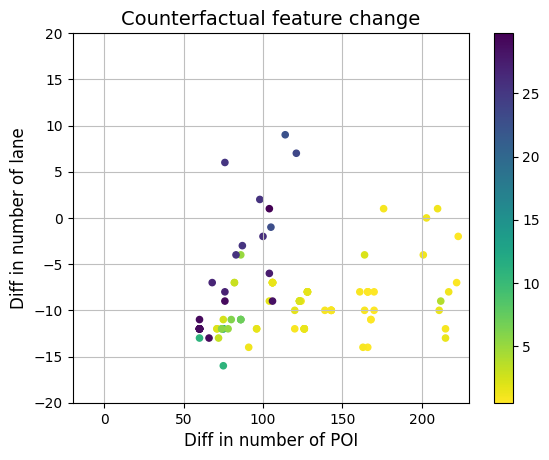}
    & \includegraphics[width=0.3\textwidth]{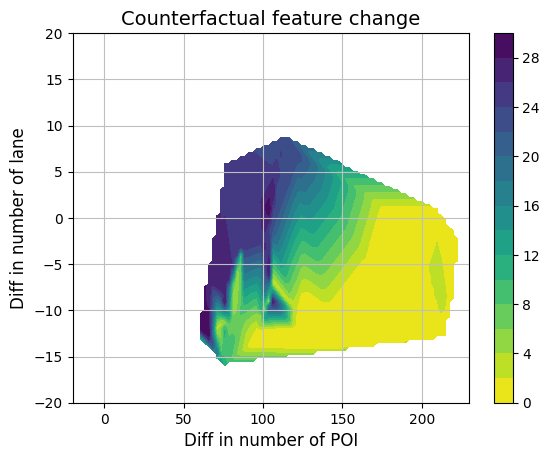}
    & \includegraphics[width=0.3\textwidth]{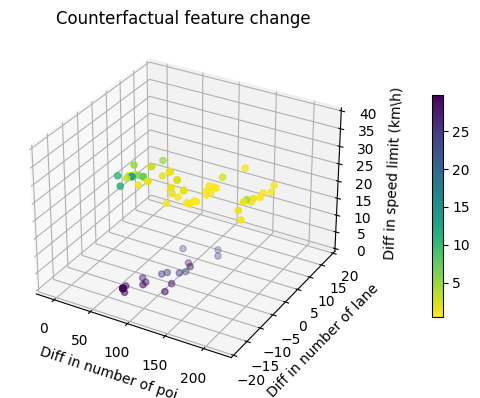}  
  \end{tabularx}
  \caption{Results for various directional constraints. Scenario A has no extra constraint; scenario B involves a decrease in the number of POIs and an increase in the number of lanes; scenario C involves an increase in the number of POIs. The ``Scatter" column displays a scatter plot of the total feature change, where the color bar represents the validity score—the brighter the color, the better the counterfactual performance. The ``Interpolation" column provides a linear interpolation based on the scatter plot data. The ``3D Scatter" column presents a 3-dimensional scatter plot incorporating total feature changes, including variations in speed limit as z-axis.}
  \label{directionalconstraint}
\end{figure}

In addition, we examine the distribution of the counterfactual explanations in terms of their total feature changes (Figure \ref{directionalconstraint}). The scatter plot visualizes the cumulative feature changes for each generated counterfactual explanation. In the 2D scatter plot, the axes represent the variations in the number of POIs and the number of lanes. Larger values on these axes signify greater differences between the counterfactual and original features. The 3D scatter plot adds a z-axis to display changes in speed limits. The color bar shows the validity score associated with each counterfactual, with a brighter color denoting a better performance of the counterfactual explanation. Detailed analyses of the three scenarios are presented below.

\subsubsection{Scenario A: No directional constraints}
In this baseline scenario, counterfactual explanations were generated from section \ref{baseline_exp}.
The scatter plot and its corresponding linear interpolation suggest that counterfactual explanations involving a greater increase in the number of POIs and a decrease in the number of lanes tend to yield superior performance, as evidenced by lower validity loss. This observation aligns well with previous findings specific to suburban roads.
The 3D scatter plot illustrates that the larger the increase in the speed limit, the better the performance of the counterfactual. 

\subsubsection{Scenario B: Decrease in POIs, Increase in Lanes}

In this scenario, the counterfactual explanations are generated with directional constraints to reduce the number of POIs and increase the number of lanes. Based on the results in Figure \ref{directionalconstraint}, regarding the change in the number of POIs for each counterfactual, the distribution range remains relatively stable. In contrast, the distribution range for the change in the number of lanes broadens, with an increasing number of counterfactuals reflecting a lane increase.
Another noteworthy observation is that when this constraint is applied, the resulting counterfactual explanations tend to be associated with brighter colors on the validity score scale, implying lower validity loss. This suggests that these constrained counterfactuals generally outperform those generated under the original, unconstrained setting.

\subsubsection{Scenario C: Increase in POIs}
City planners may, at times, wish to enhance the infrastructure surrounding roads by introducing additional amenities like parking spaces, restaurants, or gas stations. However, they often aim to do this without adversely impacting road traffic. For this scenario, the aim is to increase the number of POIs and see how it affects the predicted traffic. Consequently, large penalties were applied to counterfactual features that proposed a decrease in POIs.

The scatter plot indicates a shift in the distribution of the difference in the number of POIs for the generated counterfactual explanations. This shift leans towards a higher count, suggesting that the counterfactual explanations, under this constraint, tend to propose a greater number of POIs compared to the unconstrained baseline. Meanwhile, the distribution concerning the difference in the number of lanes remains unchanged.

\subsection{Weighting Constraints}

User experience and expertise can guide the assignment of importance to different features, effectively serving as another layer of constraint.
In this study, the target is consistently set for node B, an urban road. The aim is to improve the predicted speed between 9:00 and 10:00 to achieve a target speed of 56 km/h.
\begin{figure}[tp]
  \centering
  \renewcommand{\arraystretch}{1}
  \begin{tabularx}{0.9\linewidth}{@{}
      l
      X @{\hspace{6pt}}
      X @{\hspace{6pt}}
      X
    @{}}
    & \multicolumn{1}{c}{Scatter}
    & \multicolumn{1}{c}{Interpolation} 
    & \multicolumn{1}{c}{3D scatter} \\
    \makebox[20pt]{\raisebox{40pt}{\rotatebox[origin=c]{90}{Scenario D}}}%
    & \includegraphics[width=0.3\textwidth]{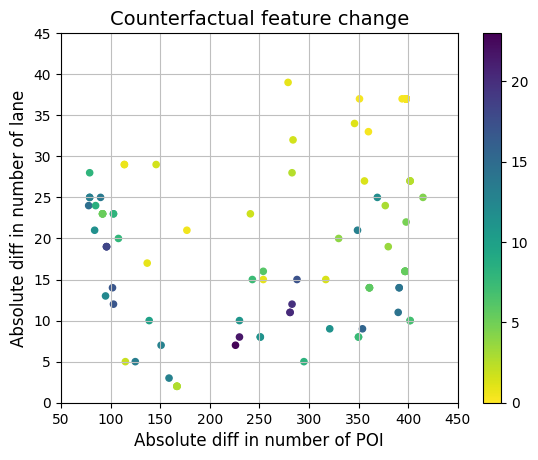}
    & \includegraphics[width=0.3\textwidth]{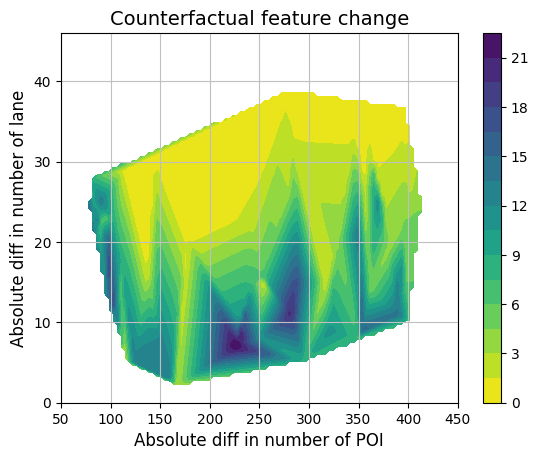} 
    & \includegraphics[width=0.3\textwidth]{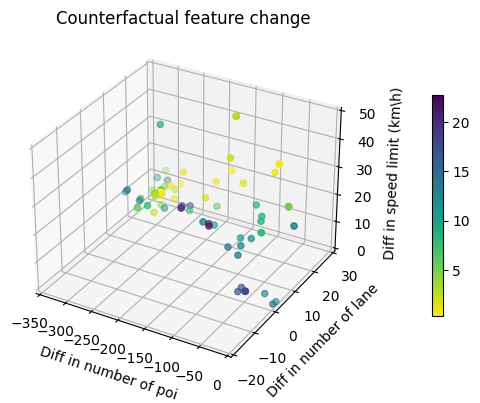}\\
    \makebox[20pt]{\raisebox{40pt}{\rotatebox[origin=c]{90}{Scenario E}}}%
    & \includegraphics[width=0.3\textwidth]{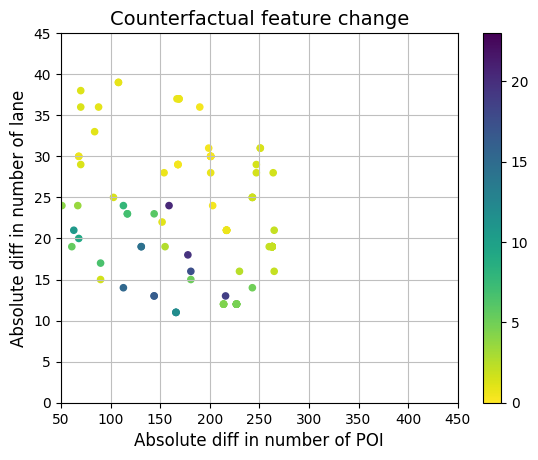}
    & \includegraphics[width=0.3\textwidth]{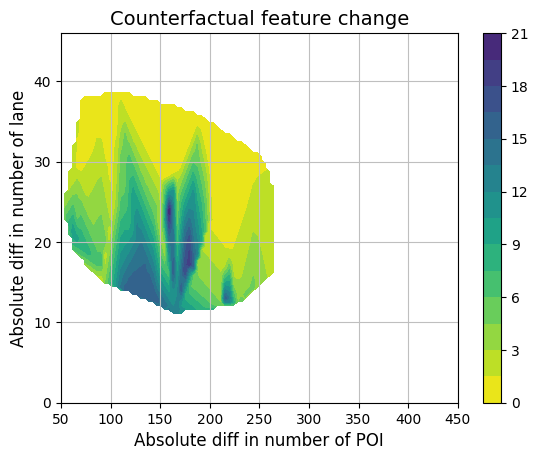} 
    & \includegraphics[width=0.3\textwidth]{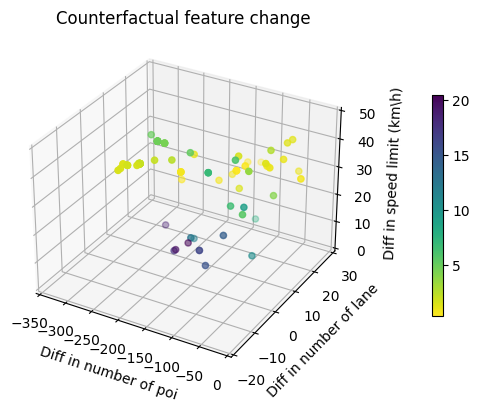}\\
    \makebox[20pt]{\raisebox{40pt}{\rotatebox[origin=c]{90}{Scenario F}}}%
    & \includegraphics[width=0.3\textwidth]{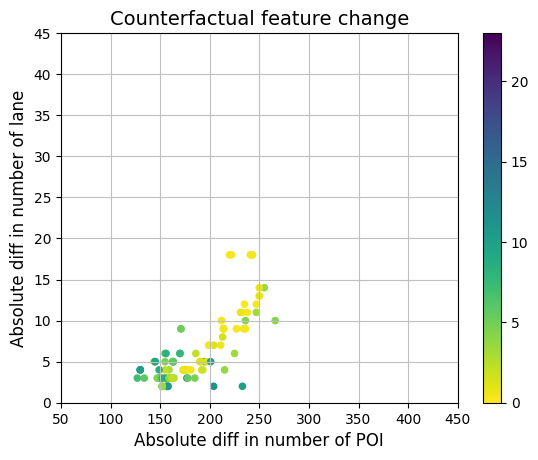}
    & \includegraphics[width=0.3\textwidth]{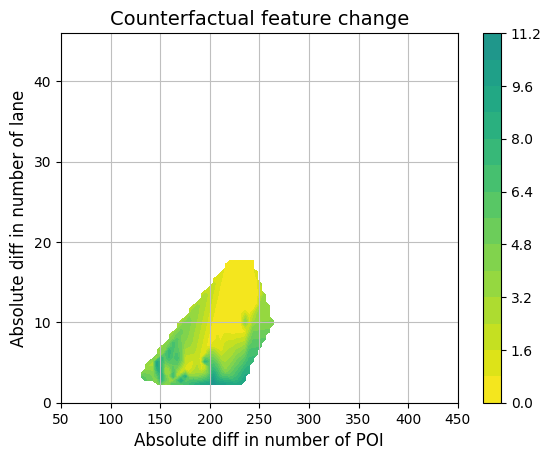} 
    & \includegraphics[width=0.3\textwidth]{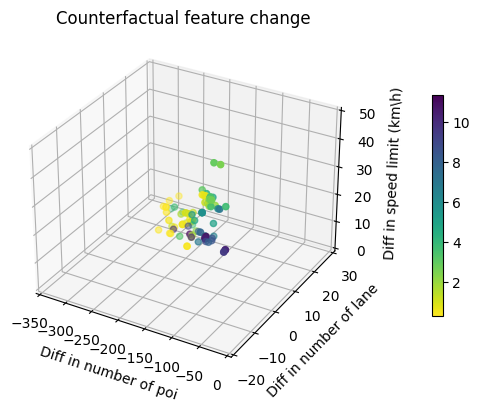}\\
    \makebox[20pt]{\raisebox{40pt}{\rotatebox[origin=c]{90}{Scenario G}}}%
    & \includegraphics[width=0.3\textwidth]{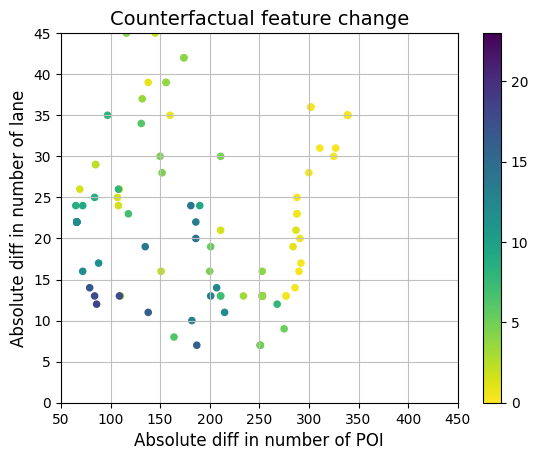}
    & \includegraphics[width=0.3\textwidth]{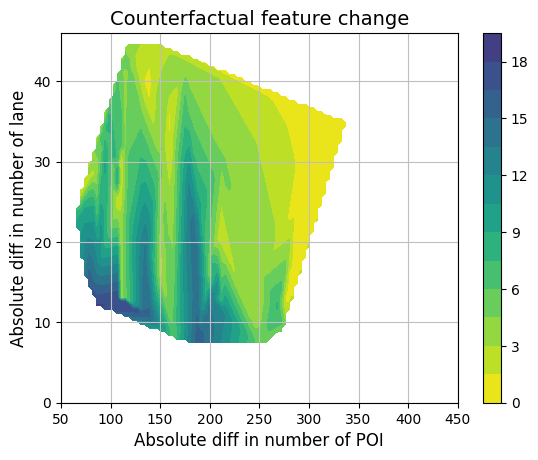} 
    & \includegraphics[width=0.3\textwidth]{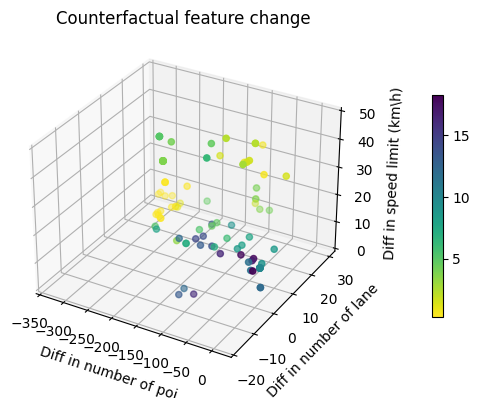}\\
    \makebox[20pt]{\raisebox{40pt}{\rotatebox[origin=c]{90}{Scenario H}}}%
    & \includegraphics[width=0.3\textwidth]{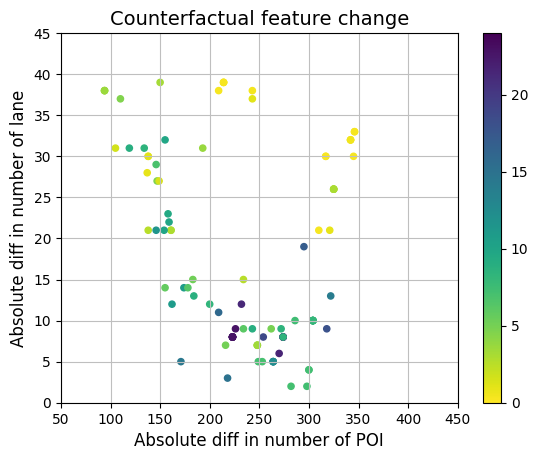}
    & \includegraphics[width=0.3\textwidth]{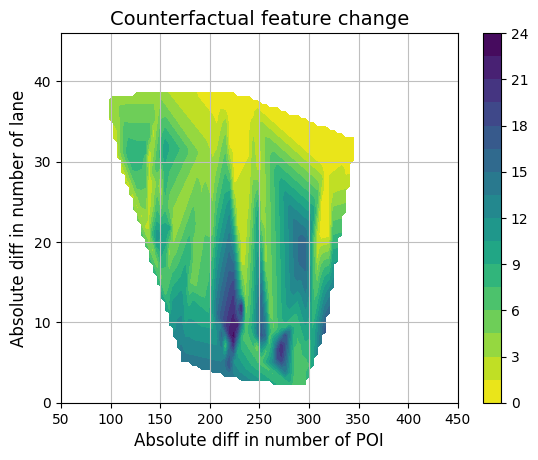} 
    & \includegraphics[width=0.3\textwidth]{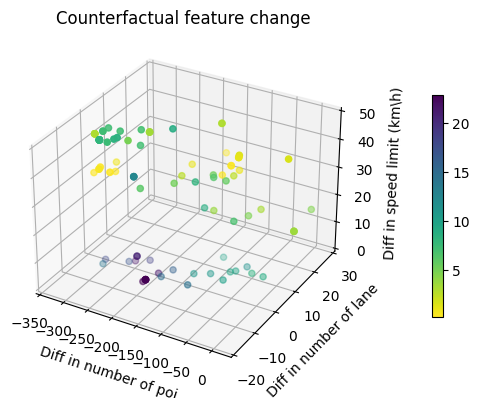}\\  
  \end{tabularx}
  \caption{Results for various weighting constraints. Scenario D has no extra constraint; scenario E preserves the number of POIs; scenario F preserves the number of Lanes; scenario G preserves both the number of POIs and the number of lanes; scenario H preserves the speed limit.}
  \label{weightingconstraint_result}
\end{figure}
Figure \ref{weightingconstraint_result} visualizes the results of the generated counterfactual explanations under different constraints.
It is worth noting that the scatter plot in Figure \ref{weightingconstraint_result} displays the absolute differences between the original and counterfactual features.

\subsubsection{Scenario D: No Weighting Constraints}
In this scenario, the counterfactual explanations generally perform better with a larger change in the number of lanes, while there is no discernible trend for the change in the number of POIs. As for the 3D scatter plot, it fails to indicate any significant correlation between variations in speed limit and the performance of the counterfactual explanations in terms of validity.

\subsubsection{Scenario E: Preserve Number of POIs}
In this configuration, we assign a higher weight to the number of POIs to discourage substantial alterations to this feature. The scatter plot and its corresponding interpolation reveal a narrower distribution range for the absolute difference between the counterfactual and original number of POIs, validating the efficacy of this weighting approach. 
Noticeably, when the changes to the number of POIs are constrained, the distribution of alterations in the number of lanes tends to cluster towards the higher end of the range.

\subsubsection{Scenario F: Preserve Number of Lanes}
In this setup, a higher weight is allocated to the number of lanes with the objective of minimizing alterations to this attribute. Both the scatter plot and the interpolation exhibit a constricted distribution range for the absolute difference between the original and counterfactual number of lanes. This outcome substantiates the effectiveness of this weighting strategy.
It is noteworthy that when modifications to the number of lanes are restricted, the distribution of changes in the number of POIs also becomes more condensed. Compared to Scenario D, the scatter points are markedly clustered towards smaller differences in both lanes and POIs' counts. 
Moreover, adding this constraint appears to enhance the overall validity performance of the counterfactuals.

\subsubsection{Scenario G: Preserve Both Number of POIs and Number of Lanes}
In this setup, significant weights are allocated to both the number of POIs and lanes, guiding the model to focus on modifying speed limits. Interestingly, the scatter plot shows that this constraint only moderately limits alterations in the number of POIs. Moreover, it does not restrain changes in lane count. With respect to counterfactual performance, more effective counterfactuals seem to be concentrated in areas showing larger differences in the number of POIs.
As for the overall performance of the set of counterfactual explanations, a decrease in validity loss suggests enhanced efficacy.

\subsubsection{Scenario H: Preserve Speed Limit}
This scenario attaches a high weight to the speed limit, directing the model to search for counterfactuals that predominantly alter other features while keeping the original speed limit intact.
Based on the scatter plot, this constraint does not yield a noticeable impact on the magnitude of changes in any specific counterfactual features. In terms of performance, higher-quality counterfactuals are more likely to be located in regions showing substantial differences in the number of lanes.

\section{Discussion}

\subsection{Impact of contextual data on traffic forecasting}

To affirm the assumption that incorporating contextual features enhances traffic forecasting, we undertook a systematic performance evaluation of models trained on various datasets.
All models underwent training across 80 epochs for a fair comparison.
The baseline model, trained exclusively on speed data, serves as the point of reference. Subsequently, we incorporated each contextual feature into the training data at a time and compared the resultant model's performance with that of the model trained using both speed and all contextual data.

\begin{table}[htb] 
    \centering
    \resizebox{\columnwidth}{!}{
    \begin{tabular} {c|ccccccccccc}
         \toprule
   Metrics& Baseline & Number of lanes & Number of POI & Speed limit &Temperature & Precipitation & Wind & Humidity & Hour of day & Day of week & Full data  \\
   \midrule
   RMSE & 10.2676 & 10.1596 & 10.2214 & 9.9265 & 10.2018& 10.2208& 10.2025 & 10.2295 & 10.2362 & 10.1841 & \textbf{9.7578}\\
   MAE & 6.8945 & 6.6427 & 6.7095 & 7.0003 & 6.6097 & 6.6260 & 6.6018& 6.7878 & 6.8190 & 6.8221 & \textbf{6.4914}\\
   Accuracy & 84.35\% & 84.50\% & 84.42\% & 84.87\% & 84.45\% & 84.42\% & 84.44\% & 84.40\% & 84.39\% & 84.47\% & \textbf{85.12\%}\\
   \(R^2\) & 0.7709& 0.7754 & 0.7727 & 0.7874 & 0.7738 & 0.7729 & 0.7737 & 0.7724 & 0.7722 & 0.7747 & \textbf{0.7931}\\
   VAR & 0.7719 & 0.7754 & 0.7727 & 0.7939 & 0.7745 & 0.7732 & 0.7744 & 0.7727 & 0.7727 & 0.7757 & \textbf{0.7940}\\
   Loss & 105.4242 & 103.2179 & 104.4773 & 98.5349 & 104.0777 & 104.4658 & 104.0910 & 104.6418 & 104.7800 & 103.7168 & \textbf{95.2140}\\

   \bottomrule
    \end{tabular}}
    \caption{Traffic forecasting model performance with different training datasets. Baseline indicates the model trained with only speed data. Full data shows the model trained with speed data and all contextual features. The ``Loss" metric presents the loss value for the test data.}
    \label{effectofcontext}
\end{table}

As illustrated in Table \ref{effectofcontext}, the comprehensive model that incorporates all contextual features demonstrates superior performance across all the evaluation metrics. It has the lowest values for RMSE, MAE, and Loss while achieving the highest scores in Accuracy, \(R^2\), and VAR.
At the same time, the model trained without any contextual data exhibited the least effective performance.
It is worth noting that although these contextual features contribute to model accuracy, their overall enhancement of predictive performance is relatively limited, resulting in a modest reduction of merely 0.4 km/h in error, which suggests their role might be less critical in terms of model training. 
However, the utility of these features is notably underscored through the application of counterfactual explanations. With CFEs, it is possible to alter the prediction results with minor changes in the input contextual features, which can tell us the importance of input features in terms of sensitivity.

\subsection{Comparison of CFEs in various spatial and temporal configurations}

\subsubsection{Impact of contextual features on highway traffic}
Counterfactual explanations generated for highway road segments failed to yield improvements in speed. This suggests that the static features investigated in this study, namely the number of POI, the number of lanes, and speed limits, do not substantially influence traffic patterns on highways within the scope of this road network.

In the case of nearby POIs, their presence appears to have negligible impact on highway speeds, as highways generally lack direct access to these facilities. Regarding the number of lanes and speed limits, isolated adjustments to these parameters on specific highway segments seem ineffective at altering overall predicted speed. This is likely because highway traffic speed at a specific time is highly dependent on near historical traffic speeds and inflow conditions; altering the static attributes of only a section of the highway would not significantly impact the overall traffic demand or the carrying capacity of the entire highway network. Therefore, it will not increase the predicted speed in this situation.

\subsubsection{Impact of contextual features on suburban road}
When aiming to increase predicted speeds on suburban road segments, counterfactual explanations suggest an increase in the number of POIs nearby. This is because the model associates road segments with a higher density of nearby POIs with lower levels of traffic congestion.

The geographical location of a suburban road appears to significantly influence its traffic patterns. For instance, suburban roads adjacent to residential neighbourhoods may experience lighter traffic but with more nearby POIs. In contrast, other suburban roads might be part of arterial routes and, despite having fewer nearby POIs, experience higher traffic volumes, leading to increased congestion or reduced speeds. It is likely the deep learning model captured these associations, therefore the CFE recommends increasing the number of nearby POIs when trying to improve predicted speeds on specific suburban roads. This alteration makes these road segments contextually similar to quieter, residential suburban roads, where lower traffic volumes and less congestion are observed.

With regard to the number of lanes, the CFE does not suggest any significant modifications, except for the case of weekday afternoons, when the original traffic is the most congested and experiences the lowest speed. During these hours, the counterfactual explanations recommend reducing the number of lanes.  
Specifically, by reducing the number of lanes at the beginning of the road segment, less traffic would be able to enter the road segment, leading to more free traffic flow and overall higher speeds.

During weekends, the CFEs did not recommend alterations to the speed limit. This suggests that speed limits are not a significant factor affecting suburban road traffic forecasting during these times.

\subsubsection{Impact of contextual features on urban road}

In contrast to the suburban road, when targeting to increase speeds on urban road segments, counterfactual explanations suggest a decrease in the number of POIs nearby. 

This discrepancy between urban and suburban roads could be interpreted in two ways. Firstly, it reflects the inherently different traffic patterns between suburban and urban settings. Secondly, it is important to note that the initial number of POIs near the studied urban road segments is already quite high. Unlike in suburban areas where an increase in POIs seems to alleviate congestion, urban roads appear to benefit from a reduction in POIs, presumably because fewer attractions would lead to less traffic. Interestingly, an exception arises during weekday afternoons, where the counterfactual explanations do not recommend a reduction in the number of POIs for urban roads. This could be because, during these peak hours, the number of POIs does not have a significant influence on the speed of traffic on urban roads.

\subsection{Effectiveness of scenario-driven counterfactual explanations}

The experimental results, obtained by incorporating various scenario constraints into the counterfactual explanation generation process, are highly promising for several reasons.

Firstly, all generated counterfactual explanations demonstrate reasonable validity and plausibility scores. This indicates that the method retains its efficacy to reach the set target even when additional constraints are applied, thereby affirming the feasibility and effectiveness of the approaches proposed in this study.

Secondly, some constraints facilitate more efficient counterfactual generation. On the one hand, the collection of generated Counterfactual Explanations generally exhibits lower validity loss, implying proper performance in aligning the predicted speeds with target speeds. On the other hand, underweighting constraints, see in Figure \ref{weightingconstraint_result}, not only do the colours in the set of CFEs become more vibrant, but the scatter points also converge within a smaller area. This indicates increased efficiency after adding the scenario constraint, as the algorithm is more adept at identifying optimal counterfactuals within a constrained search space.

In summary, the integration of user-defined prior knowledge into post-hoc explanations has proven to be valuable. This not only addresses the initial research questions posed but also has profound implications for future work in the field of Explainable AI.

\subsection{Limitations and potential work}

The use of deep learning models, coupled with Counterfactual Explanations, provides a powerful combination for uncovering complex relationships between variables. These relationships may be too subtle or intricate for humans to notice, thus highlighting the novel capabilities of explainable AI and deep learning in data analysis.

However, the efficacy of this approach is bound by certain limitations. Primarily, the model's predictive and interpretative strengths depend on the quality and diversity of the training data. Similar to many other data-driven methods, the model's generalizability may be limited by the lack of data variability, resulting in recommendations that are not broadly applicable to other cases. In the context of this study, a noteworthy limitation lies in the restricted exploration of limited road segments and contextual features. This narrow scope may influence the robustness of the generated counterfactuals and their applicability to other scenarios.

One potential avenue for mitigating these limitations involves the incorporation of domain-specific knowledge into the data-driven models. This can enhance the generalizability and reliability of the model's recommendations. 
In light of this, scenario-driven counterfactual explanations are proposed. While our work demonstrates that scenario-driven counterfactual explanations offer considerable benefits in the context of integrating prior constraints, a key question that remains is how to ensure the practical utility and broader applicability of these methods in real-world settings.

In this study, the quality of counterfactual explanations is solely evaluated based on objective metrics such as proximity and plausibility loss. We make the assumption that lower scores on these metrics indicate that implementing the counterfactual features in practice would be easier and more feasible. However, real-world applications often prove to be far more complex and challenging. To bridge this gap, future research should focus on collaborating with domain experts, such as urban planners, to gain insights into the actual challenges and constraints involved in modifying contextual settings.

\section{Conclusion}

We introduce a comprehensive framework that advances the use of counterfactual explanations in spatiotemporal prediction tasks, effectively bridging the gap between theoretical understanding of models and their practical implications for generating insights.

In this study, a deep learning-based traffic forecasting model was trained at first, using the state-of-the-art architecture, attribute augmented spatiotemporal graph convolutional networks. Subsequently, we generated diverse sets of counterfactual explanations by targeting various spatial and temporal settings.

On the one hand, by suggesting minimal alterations to input features, counterfactual explanations enhance our understanding of the model's behavior and elucidate the role of various contextual variables in deep learning-based traffic forecasting. This provides invaluable insights for AI practitioners, aiding in a deeper comprehension of what the model has learned from the data. 
More specifically, by examining a variety of spatial settings—such as suburban roads, urban roads, and highways, as well as different time slots, this study reveals that the impact of static contextual features on traffic speed is influenced by distinct spatial and temporal conditions. 
On the other hand, this study advances the field by introducing scenario-driven counterfactual explanations, which offer domain experts like urban planners insightful recommendations tailored to specific scenarios. By integrating user-defined constraints into our framework, we can provide insights that are directly applicable to a range of real-world conditions. 
Specifically, we introduce two methods for incorporating these scenario constraints: directional and weighting constraints. Both approaches effectively align the generated counterfactual explanations with users' prior knowledge and expectations, thereby making the search for optimal solutions more efficient. Importantly, we observed that some scenarios, particularly those incorporating weighting constraints, expedited the generation process and yielded more precise and useful CFEs. This is manifested through a more focused distribution of CFEs, indicating a clearer pathway for the algorithm to identify optimal counterfactual conditions.

Although this study has successfully leveraged counterfactual explanations to interpret traffic forecasting models and provided valuable insights via scenario-driven counterfactuals, several promising avenues for future research exist. Upcoming investigations could focus on:

\begin{itemize}
\item \textbf{Geographic Generalizability:} The current framework relies heavily on data from a specific geographical region. Future studies should aim to validate and adapt the model across diverse geographical settings, thereby assessing its ability to generalize the identified correlations between contextual features and traffic behaviors.

\item \textbf{Fine-Grained Feature Analysis:} While the present study broadly examines the impact of Points of Interest (POIs) on traffic dynamics, subsequent research should delve into how different categories of POIs individually influence traffic patterns.

\item \textbf{Inclusion of Dynamic Temporal Elements:} This study primarily focuses on altering static features for generating counterfactuals. Future research should expand the scope to include conducting counterfactuals on time-dependent features, potentially unveiling intricate, time-sensitive patterns that impact traffic conditions. This would entail the development of time-series counterfactual explanations, which is still an under-explored area in current literature. 

\item \textbf{Collaboration with Domain Experts:} Future work should actively involve domain experts, such as urban planners, to better incorporate real-world insights and practical constraints in the modeling process. This collaboration will improve the model's applicability and utility in decision-making processes.
\end{itemize}

\section*{Acknowledgments}
We would like to thank HERE Technologies for providing the traffic data used in this study. The study is supported by the Hasler Foundation under the project Interpretable and Robust Machine Learning for Mobility Analysis (grant number 21041). The study is partially conducted at the Future Resilient Systems program at the Singapore-ETH Centre, supported by the National Research Foundation (NRF) Singapore under its Campus for Research Excellence and Technological Enterprise (CREATE) program.

\bibliographystyle{unsrt}  
\bibliography{references}

\end{document}